\documentclass[nohyperref]{article}

\usepackage{microtype}
\usepackage{graphicx}
\usepackage{subfigure}
\usepackage{booktabs} % for professional tables
\usepackage{hyperref}

\usepackage{arxiv}

\usepackage{amsmath}
\usepackage{amssymb}
\usepackage{mathtools}
\usepackage{amsthm}

\usepackage{enumitem}
\usepackage{comment}

\usepackage[natbibapa]{apacite}
\usepackage{float}
\newcommand\numberthis{\addtocounter{equation}{1}\tag{\theequation}}

\usepackage[capitalize,noabbrev]{cleveref}

\theoremstyle{plain}

\theoremstyle{definition}

\theoremstyle{remark}

\usepackage[textsize=tiny]{todonotes}

\title{Universal Hopfield Networks: A General Framework for Single-Shot Associative Memory Models}

\begin{document}

\author{
 Beren Millidge \\
 MRC Brain Network Dynamics Unit \\
 University of Oxford \\
 \texttt{beren@millidge.name}
 \And
 Tommaso Salvatori \\
 Department of Computer Science \\
 University of Oxford \\
 \And 
 Yuhang Song \\
 Department of Computer Science \\
 University of Oxford \\
 \And
 Thomas Lukasiewicz \\
 Department of Computer Science \\
 University of Oxford \\
 \And
 Rafal Bogacz \\
 MRC Brain Network Dynamics Unit \\
    University of Oxford
    }
    
\maketitle

\begin{abstract}
    A large number of neural network models of associative memory have been proposed in the literature. These include the classical Hopfield networks (HNs), sparse distributed memories (SDMs), and more recently the modern continuous Hopfield networks (MCHNs), which possess close links with self-attention in machine learning. In this paper, we propose a general framework for understanding the operation of such memory networks as a sequence of three operations: \emph{similarity}, \emph{separation}, and \emph{projection}. We derive all these memory models as instances of our general framework with differing similarity and separation functions. We extend the mathematical framework of \citet{krotov2020large} to express general associative memory models using neural network dynamics with local computation, and derive a general energy function that is a Lyapunov function of the dynamics. Finally, using our framework, we empirically investigate the capacity of using different similarity functions for these associative memory models, beyond the dot product similarity measure, and demonstrate empirically that Euclidean or Manhattan distance similarity metrics perform substantially better in practice on many tasks, enabling a more robust retrieval and higher memory capacity than existing~models. 
\end{abstract}

\section{Introduction}

Associative, or `semantic`, memories are memory systems where data points are retrieved not by an explicit address, but by making a query to the system of approximately the same type as the data points that it stores. The system then returns the closest data point to the query according to some metric. For instance, an associative memory system, when given an image, can be used to return other `similar' images. It is often argued that the brain similarly stores and retrieves its own memories  \citep{hinton2014parallel,rolls2013mechanisms,tsodyks1995associative}, as it is a common experience to be able to recall a memory given a partial cue, e.g., recalling a song given just a few notes \citep{bonetti2021}. A large literature of neuroscience and computational theories has developed models of how such associative memory systems could be implemented in relatively biologically plausible neural network architectures~\citep{kanerva1988sparse,kanerva1992sparse,hopfield1982neural,hinton2014parallel}.

\begin{figure*}
\centering

\includegraphics[width=\textwidth]{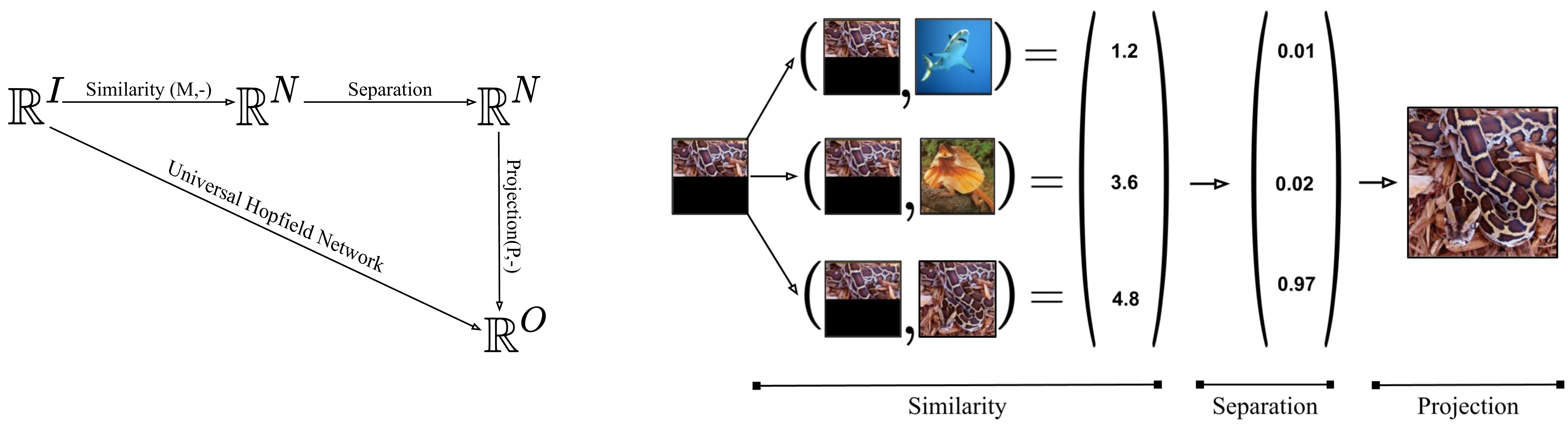}
    \vspace{-0.5cm}
  \caption{Left: Schematic of the key equations that make up the general theory of the abstract Hopfield network, which shows the factorization of a UHN into \emph{similarity, separation}, and \emph{projection}. Right: Visual representation of the factorization diagram when performing an associative memory task on three stored memories. The corrupted data point is scored against the three memories (\emph{similarity}). The difference in scores are then exaggerated (\emph{separation}), and used to retrieve a stored memory (\emph{projection}).}
  \label{fig:schematic}
  %\vspace{-0.3cm}
 \end{figure*}
 % Link to the Figure, in case you want to edit stuff: https://docs.google.com/drawings/d/1CdnrsKf_cep1Hjk70XRXcJp_7thtZHGRqEqNjn7mUuY/edit?usp=sharing

Two classical and influential models are the Hopfield network (HN) \citep{hopfield1982neural,hopfield1984neurons} and the sparse distributed memory (SDM) \citep{kanerva1988sparse, kanerva1992sparse,jaeckel1989alternative}. More recently, %these classical models 
they have been generalized to the modern-continuous Hopfield network (MCHN) \citep{ramsauer2020hopfield} and the modern-continuous sparse distributed memory (MCSDM) \citep{bricken2021attention}, which have substantially improved performance, close relationships with transformer attention, and can handle continuous~inputs. 

Here, we propose a unified framework that encompasses all  these models as simple instantiations of a more general framework, which we call \emph{universal Hopfield networks (UHNs)}.  Mathematically, the UHN can be described as a function $\text{UHN}\colon \mathbb{R}^I \rightarrow \mathbb{R}^O$ mapping a vector in an input space of dimension $I$ to a vector in an output space of dimension $O$, with two additional inputs of a memory matrix $M$ of size $N \times I$, consisting of a set of $N$ stored patterns, and a projection matrix $P$ of size $O \times N$, consisting of a potentially different set of stored patterns with dimension $O$ for heteroassociation. The dimensionality of the input and output patterns are allowed to differ to enable heteroassociative memories to be described in the same framework. For autoassociative memories $I = O$. The UHN function can be factorized into a sequence of three operations:  \emph{similarity}, \emph{separation}, and \emph{projection}, illustrated in Figure \ref{fig:schematic}. First, \emph{similarity} matching between the query and a stored set of memory vectors to produce a vector of similarity scores. Second, \emph{separation} to numerically magnify small differences in original similarity scores into large differences in the output scores so as to increase the relative separation of the scores, and finally \emph{projection}, which multiplies the resulting set of output similarity scores with a projection matrix, and constructs an output based essentially on a list of stored data points in the memory\footnote{For heteroassociative memory models, a separate projection memory is used containing the outputs associated with each input.} weighted by the output similarity scores, so that the network's output is most influenced by memories similar to the query vector. The main contributions of this paper are briefly as follows.%\vspace*{-1ex} 
\begin{itemize}[topsep=0pt,parsep=0pt,partopsep=0pt]
    \item We define a general framework of \emph{universal Hopfield networks}, which clarifies the core computation underlying single-shot associative memory models.
    %\vspace{-0.2cm}
    \item We demonstrate how existing models in the literature are special cases of this general framework, which can be expressed as an extension of the energy-based model proposed by \citet{krotov2020large}.
    %\vspace{-0.2cm}
    \item We demonstrate that our framework allows straightforward generalizations to define novel associative memory networks with superior capacity and robustness to MCHNs by using different similarity functions.
\end{itemize}

It is also important to draw a distinction between feedforward and iterative associative memory models. In feedforward models, memory retrieval is performed through a fixed computation mapping a query to its retrieved output. Examples of feedforward associative memory models include the DAM \citep{krotov2016dense}, the MCHN \citep{ramsauer2020hopfield}, and the MCSDM \citep{bricken2021attention}, which effectively instantiate 2-layer MLPs and perform retrieval as a feedforward pass through the network. Conversely, iterative associative memory models retrieve memories by either iterating over neurons \citep{hopfield1982neural,demircigil2017model} or else iterative over multiple forward passes of the network by feeding back the output into the network as a new input. It has been shown empirically that standard  autoencoder networks \citep{radhakrishnan2018memorization,radhakrishnan2020overparameterized,jiang2020associative} and predictive coding networks \citep{salvatori2021associative} can store memories as fixed points of these dynamics. In Section 3 and our experiments, we primarily investigate feedforward associative memories, while in Section 4 we derive a general framework and energy function that can support both feedforward and iterative associative memory models.

The rest of this paper is organized as follows. In Section 2, we define the mathematical framework of \emph{universal Hopfield networks}. In Section 3, we show how existing models can be derived as special cases of our framework. In Section 4, we extend the neural model of \citet{krotov2020large} to define an energy function and associated neural dynamics for the UHN. In Section 5, we show that our framework enables generalization to novel similarity and separation functions, which result in higher capacity and more robust networks, while experiments on the separation functions empirically confirm theoretical results regarding the capacities of associative memory models.
\section{Universal Hopfield Networks (UHNs)}
A single-shot associative memory can be interpreted as a function that takes an input vector $q$ (ideally, a corrupted version of a data point already in memory) and outputs a vector corresponding to the closest stored data point. Mathematically, our framework argues that every feedforward associative memory in the literature admits the following factorization, which defines an abstract and general \emph{universal Hopfield network (UHN)}:%
\begin{align}
    z = \underbrace{P}_{\text{Projection}} \cdot \underbrace{\text{sep}}_{\text{Separation}}(\underbrace{\text{sim}(M,q)}_{\text{Similarity}})\,,
    \label{core_eq}
\end{align}
where $z$ is the $O \times 1$ output vector of the memory system, $P$ is a projection matrix of dimension $O \times N$, $sep$ is the separation function, $sim$ is the similarity function, $M$ is an $N \times I$ matrix of stored memories or data points, and $q$ is the query vector of dimension $I \times 1$. 

\begin{table*}
\begin{center}

\medskip 
\begin{tabular}{ p{7cm}p{4cm}p{4cm} }
% \hline
% \multicolumn{3}{|c|}{\textbf{ASSOCIATIVE MEMORY NETWORKS}} \\
 \toprule
 \textbf{Memory Network} & \textbf{Similarity Function} & \textbf{Separation Function}  \\
 \midrule
 (Classical) Hopfield Network  (HN) & Dot Product    & Identity \\
 Sparse Distributed Memory (SDM) &   Hamming Distance  & Threshold \\
 Dense Associative Memory (DAM) & Dot Product & Polynomial  \\
 Modern Continuous Hopfield Network (MCHN)   & Dot Product & Softmax  \\
 \bottomrule
\end{tabular}
\end{center}
\vspace{-0.3cm}
\caption{Associative memory models.}
\vspace{-0.5cm}
\label{summary_table}
\end{table*}
%\end{center}
The intuition behind this computation is that, given an input query, we first want to rank how similar this query is to all the other stored memories. This is achieved by the \emph{similarity} function, which outputs a vector of similarity scores between each data point held in the memory and the query. Given these similarity scores, since we will ultimately be retrieving based on a linear combination of the patterns stored in the projection matrix, weighted by their similarity scores, and we ultimately only want to produce one clear output pattern without interference from the other patterns, then we need a way to emphasize the top score and de-emphasize the rest. This is achieved by the \emph{separation} function. It is well known that separation functions of higher polynomial degrees lead to capacity increases of the order of $C \propto N^{n-1}$, where $N$ is the number of visible (input) neurons, and $n$ is the order of the polynomial \citep{chen1986high,horn1988capacities,baldi1987number,abbott1987storage,caputo2002storage,krotov2016dense}, while exponential separation functions (such as the softmax) lead to exponential memory capacity \citep{demircigil2017model,ramsauer2020hopfield}. Taking this further, it is clear to see that simply using a \emph{max} separation function leads to a theoretically unbounded capacity in terms of the dimension of the query vector, since then presenting an already stored pattern as a query will always return itself as a memory. However, the `attractors' in such a network grow increasingly small so that, in practice, the real bound on performance is not the capacity but rather the ability of the similarity function to distinguish between the query and various possible stored patterns --- a pattern that is clear already with the `exponential' capacity MCHN network, which despite its theoretical exponential capacity often performs relatively poorly at retrieval in practice with corrupted or noisy queries. Finally, the \emph{projection} matrix takes the vector of separated similarity scores and maps it to the correct output expected of the network.

Importantly, Equation \ref{core_eq} can be interpreted as a feedforward pass through an artificial neural network with a single hidden layer, where the activation function of the first layer is the separation function, and the activation function of the output is linear or else is some post-processing function such as binarization (as in the classical HN). Interpretations of memory networks in this way have been independently proposed by \citet{kanerva1988sparse} for SDM and recently by \citet{krotov2021hierarchical} for the MCHN \citep{ramsauer2020hopfield}. Furthermore, coming from the other direction, recent work has also begun to suggest that standard 2-layer multi-layer perceptrons (MLPs) may naturally tend to function as associative memory models in practice. For instance, \citet{geva2020transformer} show that the feedforward layers of the transformer appear to serve as key-value memories \citep{geva2020transformer}, and it has been suggested that these feedforward layers can be replaced with simply persistent memory vectors \citep{sukhbaatar2019augmenting}.

%It is also important to note that our framework only applies to \emph{single-shot} associative memory models. However, there is an additional class of memory models that we call \emph{iterative} memory models, which either iterate over individual neurons in the memory model \citep{hopfield1982neural,demircigil2017model} or else iterate over multiple forward passes of a standard deep artificial neural network autoencoder. By iterating the network such that its prediction is fed back into itself as an input multiple times, it has been shown \citep{radhakrishnan2018memorization,radhakrishnan2020overparameterized,jiang2020associative} that the network can store memorized training examples at the fixed points of these dynamics. \citet{salvatori2021associative} showed that this can be extended beyond deep networks trained with backpropagation to predictive coding networks, suggesting that this may be a more general feature of deep networks regardless of how they are trained.

\section{Instances of Universal Hopfield Networks}

Now that we have defined our universal Hopfield network (UHN), we shall show how the currently existing main associative memory models can be derived as specific instances of the UHN. The equivalences are summarized in Table \ref{summary_table}.

\subsection{Hopfield Networks}
% quick history on hopfield networks
Hopfield networks (HNs) \citep{hopfield1982neural,hopfield1984neurons} consist of a single neural network layer that stores an array of binary memories $M = [m_1, m_2, \dots, m_N]$, where $M$ is an $N \times I$ matrix, and $I$ is the dimension of each memory vector, and $N$ is the number of memories stored. The memory arrays are then stored in a synaptic weight matrix $W = MM^T$. Memories are retrieved by fixing the input neurons to a query pattern $q$, which is a binary vector of length $I$. While the original HN of \citet{hopfield1984neurons} iteratively minimized the energy function over individual neurons, we here describe the `feedforward' Hopfield networks described in \citep{krotov2016dense, little1974existence}, which retrieve memories by performing a forward pass through the network to compute an output $z = \mathrm{sign}(W \cdot q)$, where $sign$ is the sign function, and $z$ is the retrieved pattern and is also a binary vector of length $I$ (since the HN is autoassociative). This process can be repeated if necessary to further minimize the energy by feeding in the reconstructed output again to the network as its input. This network can be interpreted as minimizing a `Hopfield energy function', which is equivalent to the energy function of an Ising spin-glass model \citep{kirkpatrick1978infinite, keeler1988comparison}. To show that the HN is an example of a UHN, first recall that the synaptic weight matrix in the HN is defined not as the stored pattern matrix but as the outer product $W = M M^T$. By substituting this into the HN update rule, we obtain $z = \mathrm{sign}((M M^T) \cdot q) = \mathrm{sign}(M \mathcal{I} (M^T  \cdot q))$, where we use $\mathcal{I}$ to denote the identity function. Thus, we can understand the HN within our framework as using a dot-product similarity function and an identity separation function (which is the cause of the HN's relatively poor storage capacity). The sign function plays no part in memory retrieval and simply binarizes the network's output.

\subsection{Sparse Distributed Memories}
% subsection sdm
Sparse distributed memories \citep{kanerva1988sparse,kanerva1992sparse} (SDM) are designed to heteroassociate long binary vectors. The network consists of two matrices --- an `Address' matrix and a `Pattern' matrix. Memories are thought of as being stored in a data-type with both an `Address' and a `Pattern' pointer. To retrieve a memory, a query vector is compared against all stored addresses in the Address matrix, and the binary Hamming distance between the query and all addresses is computed. Then, a certain number of addresses are activated that  are below a threshold Hamming distance from the query. The memory is retrieved by summing the pattern pointers for all the addresses activated by the query. The `read' phase of the SDM \citep{kanerva1988sparse} can be written mathematically as $P \cdot \text{thresh}(d(M,q))$, where $d$ is the Hamming distance function, and $thresh$ is a threshold function that returns $1$ if the Hamming distance is greater than some threshold, and $0$ otherwise. Here, it is clear that the SDM can be naturally understood using our framework with similarity function $d$ (Hamming distance) and separation function $thresh$, which implements a top-k operation to cut out poor matches.

\subsection{Dense Associative Memories and Modern Continuous Hopfield Networks}

In recent years, the capabilities of both of these classical memory models have been substantially improved, and a number of new Hopfield architectures have been developed based on the dense associative memory (DAM) initially proposed by \citet{krotov2016dense} and extended by \citet{demircigil2017model}. Specifically, \citet{krotov2016dense} argued for generalizing the standard Hopfield energy function ($E = q^T W q + q^T b$) (where $b$ is an $I \times 1$ bias vector to convert between binary and bipolar representations) to an arbitrary function of $q$ and $W$: $E = F(W \cdot q)$ and showed that as $F$ becomes a polynomial of increasing order, the memory storage capacity of the network increases as $C \propto I^{n-1}$,  where $I$ is the number of visible neurons, and $n$ is the order of the polynomial. \citet{demircigil2017model} extended this argument to exponential energy functions of the form $E = \sigma(W \cdot q)$, where $\sigma(x)$ is the softmax function, and showed that the resulting networks have \emph{exponential} storage capacity. Then,  \citet{ramsauer2020hopfield} generalized these networks to continuous (instead of binary) inputs to derive the modern continuous Hopfield network (MCHN). The MCHN uses the energy function $E = q^T q + \text{logsumexp}(Wq)$, which can be minimized with the convex-concave procedure \citep{yuille2003concave}, giving the update rule $z = W^T \sigma(Wq)$, which enables exponential capacity, memory retrieval in a single step, and is extremely similar to the feedforward pass of a self-attention unit $z = V \sigma(K Q)$ with `Query Matrix' $Q$, `Key Matrix' $K$, and `Value Matrix' $V$, where we can associate $Q = q$, $K = W$, and $V = W$ \citep{bahdanau2014neural,vaswani2017attention, devlin2018bert,brown2020language,radford2019language}. Lastly, \citet{krotov2020large} presented a unified set of neural dynamics that can reproduce the original HN, the polynomial interaction functions of \citep{krotov2016dense}, and the exponential Hopfield network of \citep{demircigil2017model,ramsauer2020hopfield}, using only local computations, and which \citet{tang2021remark} have shown also to be related to the spherical normalization dynamics in the recent \mbox{MLP-mixer  \citep{tolstikhin2021mlp}.} 

\subsection{Continuous Sparse Distributed Memories}
% is hamming distance dot product mod 2?
Recent work has also uncovered a close link between SDMs and transformer attention \citep{bricken2021attention}. Recall that the SDM read rule can be expressed as $P \cdot \mathit{thresh}(d(A,q))$, where $\mathit{thresh}$ is a threshold function, $A$ is an $M \times N$ matrix of addresses, $P$ is a $K \times O$ matrix mapping each stored data point to its associated pattern, and $d$ is the Hamming distance between each of the stored addresses in $A$ and the query pattern $q$. \citet{bricken2021attention} first generalized SDM from binary vectors to the `continuous SDM`, where $P$, $A$, and $q$ contain real values instead of bits. Then, they replaced the Hamming distance (which only applies to binary vectors) with the dot product, using the argument that the Hamming distance is the dot product (mod 2) of binary vectors, and thus that the dot product is the natural generalization of the Hamming distance to continuous variables. Finally, they noted that the decay of the number of addresses that are not cutoff by the threshold function decreases approximately exponentially as the Hamming distance threshold decreases. The mathematical reason for this is that the distribution of addresses within a given Hamming distance of a query is a binomial distribution, which can be well approximated with a Gaussian at large $N$, and the tails of a Gaussian distribution decay exponentially. This approximately exponential decay in the number of addresses passing the threshold allows us to heuristically replace the threshold function with an exponential function, resulting in the following approximate update rule for the `continuous SDM` model $z = P \sigma(Aq)$, which is closely related to the self-attention update rule and is identical to the rule for the MCHN.

%\paragraph{Modern Continuous Hopfield Network:} The update rule for the MCHN \citep{ramsauer2020hopfield} is $q^* = X \sigma(X^T q)$ where $\sigma$ is a softmax function, $X$ is a $Q \times Q$ memory matrix and $q$ is a $Q \times 1$ query vector. In our framework the MCHN can be described as having a similarity function as the dot product $\text{sim}(M,q) = Mq$ and a softmax separation function.
%\paragraph{Sparse Distributed Memory:} The `read' phase of the SDM \citep{kanerva1988sparse} can be written mathematically as, $P \cdot \text{thresh}(d(M,q))$ where $d$ is the Hamming distance function and $thresh$ is a threshold function which returns $1$ if the hamming distance is greater than some threshold and $0$ otherwise. Here it is clear that the SDM can be naturally understood using our framework with similarity function $d$ (Hamming Distance) and separation function $thresh$ which implements a top-k operation to cut out poor matches.
%\paragraph{Hopfield Network:} The update rule for the classical Hopfield Network \citep{hopfield1982neural} is $q^* = \mathrm{sign}(W \cdot q)$ where $sign$ is the sign function. Despite looking subtly different, it is straightforward to show that the HN is also explainable within our framework. 
% could add energies to this table
%\begin{center}

\subsection{Auto- and Heteroassociative Memories}
\label{main_text_heteroassociative}
Our framework also provides a simple explanation of the difference between autoassociative memories (which map a corrupted version of a memory to itself) and heteroassociative memories (which map some input memory to some other memory type, potentially allowing for memory chains and sequence retrieval): namely, that autoassociative memories set the projection matrix $P$ equal to the memory matrix $M$, i.e., one recalls the memories used for similarity matching, while heteroassociative memory networks set the projection matrix equal to the associated heteroassociated memory. It is thus clear to see why the HN and MCHN networks are autoassociative, and how to convert them to heteroassociative memory networks. Namely, for the MCHN, set the update rule to $z = P \sigma(M.q)$, and for the HN set the weight matrix $W = PM^T$. Demonstrations of these novel heteroassociative HNs and MCHNs are given in Appendix B. Interestingly, the heteroassociative MCHN update rule is equivalent to the self-attention update rule found in transformer networks \citep{vaswani2017attention}, and thus suggests that the fundamental operation performed by transformer networks is heteroassociation of inputs (the queries) and memories (the keys) with other memories (the values).
%\vspace{-0.2cm}

\section{Neural Dynamics}
In this section, extending the work of \citet{krotov2020large}, we present an abstract energy function for the UHN and set of neural dynamics that minimize it, which can be specialized to reproduce any of the associative memory models in the literature. By framing associative memory models in terms of an energy function, we can describe the operation of both iterative and feedforward associative memory models, as well as mathematically investigate the properties of the fixed points that they use as memories. We define a general neural implementation and energy function for our abstract associative memory model that uses only local interactions. In this model, there are two types of `neurons': `value neurons' $v$ and `memory neurons' $h$. We can think of the `value neurons' $v$ being initialized to the query pattern $q$ such that $v^{t=0} = q$, and then updated to produce the output pattern $z$. This is because the UHN effectively implements a two-layer artificial neural network where the value neurons are the input layer and the memory neurons are the hidden layer. The memory and value neurons are interconnected by the memory matrix $M$. The neural activities $v$ and $h$ are also passed through two activation functions $g$ and $f$ such that $f = f(h)$ and $g = g(v)$. The network has the following recurrent neural dynamics:
\begin{align}
   \tau_v \frac{dv_i}{dt} &= -\frac{\partial E}{\partial v_i} = \sum\nolimits_j \frac{\partial \text{sim}(M_{i,j},v_i)}{\partial v_i} f_i - v_i g'(v_i) \\
    \tau_h \frac{d h_i}{dt} &= -\frac{\partial E}{\partial h_i} = f'(h_i) \big[ \sum\nolimits_j \text{sim}(M_{i,j},v_i) - h_i \big]\,, \nonumber
\end{align}
where $\tau_v$ and $\tau_h$ are time-constants of the dynamics. These dynamics can be derived from the energy function:
\begin{align*}
    E(M,v,h) &= \big[ \sum\nolimits_i (v_i \,{-}\, I_i)g_i   \,{-}\, \mathcal{L}_v \big] \, + \\ & \hspace*{-8ex}\big[ \sum\nolimits_i f_i h_i \,{-}\, \mathcal{L}_h \big] - \sum\nolimits_i \sum\nolimits_j f_i \text{sim}(M_{i,j},v_i)\,, \numberthis
\end{align*}

where we define the `Lagrangian' functions $\mathcal{L}_v$ and $\mathcal{L}_h$ such that their derivatives are equal to the activation functions $g = \frac{\partial \mathcal{L}_v}{\partial v}$ and $f = \frac{\partial \mathcal{L}_h}{\partial h}$. The energy function is defined such that it only includes second-order interactions between the value and memory neurons in the third term, while the first two terms in square brackets only involve single sets of neurons. In Appendix A, we show that the energy function is a Lyapunov function of the dynamics, i.e., it always decreases over time, as long as the Hessian of the activation functions $f$ and $g$ are positive definite. To obtain the dynamics equations (2) and (3), we first define function $f$ to be the separation function $f(h) = \text{sep}(h)$ such that $\mathcal{L}_h = \int dh \text{sep}(h)$ and denote $f'(h) = \frac{\partial f(h)}{\partial h}$ as the derivative of $f$. We also set $\mathcal{L}_v = \frac{1}{2} \sum_i v_i^2$, which implies $g_i = \frac{\mathcal{L}_v}{\partial v_i} = v_i$. A further simplication of the energy and dynamics occurs if we assume that $\tau_h$ is small, and thus the dynamics of the hidden neurons are fast compared to the value neurons such that we can safely assume that these dynamics have converged. This allows us to write $h^* = \sum_j \text{sim}(M_{i,j},v_i)$, since when setting $\frac{dh_i}{dt} = 0$, we can cancel the $f'(h_i)$ terms as long as $f'(h) \neq 0$, which is true for all separation functions we consider in this paper except the $\text{max}$ function, which is therefore heuristic. This gives us the simpler and intuitive energy function:
\begin{align*}
    E &= \sum\nolimits_i v_i^2 - \frac{1}{2}\sum\nolimits_i v_i^2 + \sum\nolimits_i f_i \sum\nolimits_j \text{sim}(M_{i,j} v_i) \\ &- \mathcal{L}_h - \sum\nolimits_i \sum\nolimits_j  f_i \text{sim}(M_{i,j},v_i) \\
    &= \sum\nolimits_i \frac{1}{2}  v_i^2 - \int \text{sep}(\sum\nolimits_j \text{sim}(M_{i,j},v_i))\,, \numberthis
\end{align*}
where the integral is over the input to the separation function. It is now straightforward to derive the HN and MCHN. To do so, we set $\text{sim}(M,v) = Mv$ and $\text{sep}(x) = x$ for the HN, and $\text{sep}(x) = \frac{e^{x}}{\sum e^x}$ for the MCHN. Following \citet{krotov2020large}, for the MCHN, we can derive a single step of the dynamics by taking the gradient of the energy:
\begin{align*}
    E &= \sum\nolimits_i \frac{1}{2}  v_i^2 -  \log \sum\nolimits_j  e^{(\text{sim}(M_{i,j},v_i))} \numberthis \\
   \tau_v\frac{dv_i}{dt} &=-\frac{\partial E}{\partial v_i} = - v_i + \frac{e^{\sum\nolimits_j \text{sim}(M_{i,j}, v_i)}}{\sum\nolimits_i e^{\sum\nolimits_j \text{sim}(M_{i,j}, v_i)}} \frac{\partial \text{sim}(M_{i,j} v_i)}{\partial v_i}  .
\end{align*}
If we then perform an Euler discretization of these dynamics and set $\Delta t = \tau_v = 1$, then we obtain the following update step:
\begin{align*}
    v^{t+1}_i = M \sigma(Mv_i^t)\,, \numberthis
\end{align*}
where $\sigma(x) = \frac{e^x}{\sum e^x}$ is the softmax function by using the fact that the MCHN uses the dot product similarity function $\text{sim}(M,v) = Mv$. It was proven in \citep{ramsauer2020hopfield} that this update converges to a local minimum in a single step. This thus derives the MCHN update
\begin{align}
    v^* = M \sigma(M q),
\end{align}
since $v^{t=0} = q$ as the visible neurons are initialized to the input query and equilibrium occurs after a single step.  Similarly, to derive the HN, we set the separation function to the identity ($\text{sep}(x) = x$) and similarity function to the dot product and using the fact that $\int d(Mv) \text{sep}(\text{sim}(Mv)) = \int d(Mv) Mv = \frac{1}{2}(Mv)^2$, thus resulting in the energy function and equilibrium update rule:
\begin{align*}
E &= \sum\nolimits_i \frac{1}{2} v_i^2 - \frac{1}{2} \sum\nolimits_i \sum\nolimits_j v_i M_{i,j}M_{i,j}^T v_i  \\
\tau_v\frac{dv_i}{dt} &= -\frac{\partial E}{\partial v_i} = -v_i + \sum\nolimits_i \sum\nolimits_j M_{i,j} M_{i,j}^T v_i  \,, \numberthis
\end{align*}
where, again, if we perform the Euler discretization, we obtain the following update step:
\begin{align*}
    v^{t+1} = MM^T v^t ,\numberthis
\end{align*}
which, with a final normalizing $sign$ function to binarize the output reconstruction, is identical to the HN update rule. We thus see that using this abstract energy function, we can derive a Lyapunov energy function and associated local neural dynamics for any associative memory model that fits within our framework. Moreover, our framework also describes iterative associative memory models if these inference dynamics are integrated over multiple steps instead of converging in a single step. 

\begin{figure*}
    \centering
    \includegraphics[scale=0.5]{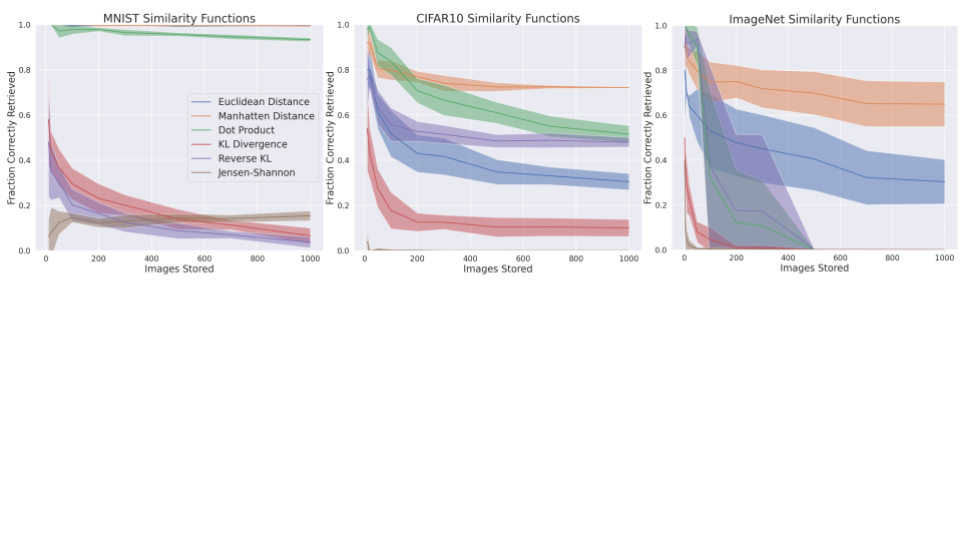}
    \vspace{-5cm}
    \caption{Capacity of the associative memory networks with different similarity functions, as measured by increasing the number of stored images. The capacity is measured as the fraction of correct retrievals. To test retrieval, the top-half of the image was masked with all zeros (this is equivalent to a fraction masked of $0.5$ in Figure \ref{mask_noise_fig}) and was then presented as the query vector for the network. A retrieval was determined to be correct if the summed squared difference between all pixels in the retrieved image and the true reconstruction was less than a threshold $T$, which was set at $50$. The queries were presented as the stored images corrupted with independent Gaussian noise with a variance of $0.5$. Mean retrievals over 10 runs with different sets of memories images. Error bars are computed as the standard deviations of the correct retrievals of the 10 runs. A softmax separation function was used with a $\beta$ parameter of 100.}
    \label{capacity_graph_figure}
\end{figure*}
%\vspace{-0.3cm}
\section{Experiments}
Our general framework allows us to define an abstract associative memory model with arbitrary similarity and separation functions, as well as a set of neural dynamics and associated energy function for that model. A natural question is whether we can use this abstract framework to derive more performant associative memory models by using different similarity and separation functions. In this section, we empirically test a wide range of potential separation and similarity functions on associative memory retrieval tasks. We find similarity functions such as the Manhattan (or absolute or $l1$ \text{norm}) distance metric perform substantially better than the dot product distance used in the MCHN across a datasets and is more robust to input distortion. We define novel associative memory models with state-of-the-art performance, which can scale beyond that considered previously in the literature, especially on the Tiny ImageNet dataset. In Appendix E, we discuss the detailed numerical normalizations and other implementation details that are necessary to achieve a good performance in practice.

\subsection{Capacity under Different Similarity Functions}
We investigate the capacity of the associative memory models to increasing numbers of stored memories on a suite of potential similarity functions. The similarity and separation functions tested are defined in Appendix D. We tested the retrieval capacity on three image datasets: MNIST, CIFAR10, and Tiny ImageNet. All images were normalized such that all pixel values lay between $0$ and $1$. Before presenting the images to the network as queries,  they were flattened into a single vector. When masking the images, the masked out pixels were set to $0$. When adding Gaussian noise to the images, we clipped the pixel values after noise was added to maintain all values between $0$ and $1$. 

%\begin{figure}[H]
%\centering
%    \begin{subfigure}
%        \centering
%        \includegraphics[width=.3\textwidth]{figures/mnist_N_runs_capacity_graph.jpg}
%        \caption{MNIST}
%    \end{subfigure}
%        \begin{subfigure}
%        \centering
%        \includegraphics[width=.3\textwidth]{figures/N_runs_capacity_graph.jpg}
%        \caption{Cifar10}
%    \end{subfigure}
 %       \begin{subfigure}
 %       \centering
%        \includegraphics[width=.3\textwidth]{figures/tiny_2_N_runs_capacity_graph.jpg}
 %       \caption{Tiny Imagenet}
%    \end{subfigure}
 %   \caption{Capacity of the associative memory networks with different similarity functions as measured by increasing the number of stored images. The capacity is measured as the fraction of correct retrievals. To test retrieval, the top-half of the image was masked with all zeros and was then presented as the query vector for the network. Retrieval was determined to be correct if the summed squared difference between all pixels in the retrieved image and the true reconstruction was less than a threshold $T$ which was set at $50$. The queries were presented as the stored images corrupted with independent Gaussian noise with a variance of $0.5$. Mean retrievals over 10 runs with different sets of memories images. Error bars are computed as the standard deviations of the correct retrievals of the 10 runs. A softmax separation function was used with a $\beta$ parameter of 100.}
 %   \label{capacity_graph_figure}
%\end{figure}

From Figure \ref{capacity_graph_figure}, we see that the similarity function has a large effect on the memory capacity of the associative memory models. Empirically, we see very robustly that the highest performing and highest capacity similarity function is the Manhattan distance $\text{sim}(M,q) = \text{abs}(M-q)$, where the subtraction is taken over rows of the memory matrix. Moreover, the superiority of the Manhattan distance as a similarity function appears to grow with the complexity of the dataset. It is roughly equivalent to the Euclidean and dot product on MNIST, slightly better on CIFAR10 and substantially better on Tiny ImageNet. The Euclidean distance also performs very well across image datasets. Other potential measures such as the KL divergence, Jensen-Shannon distance, and reverse KL perform substantially worse than simple Euclidean, dot-product, and Manhattan distance measures. The dot product metric used in the MHCN also performs very well, although it must be carefully normalized (see Appendix E). Interestingly, we see stable levels of performance for increasing capacity for a given similarity function across a wide range of memory capacities.

The similarity functions are so important, because they are the fundamental method by which the abstract associative memory model can perform ranking and matching of the query to memory vectors. An ideal similarity function would preserve a high similarity across semantically non-meaningful transformations of the query vectors (i.e., insensitive to random noise, perturbations, and masking of parts of the image), while returning a low similarity for transformed queries originating from other memory vectors. An interesting idea is that, while thus far we have used simple similarity functions such as the dot product and the Euclidean distance, it is possible to define smarter distance metrics native to certain data types, which should be expected to give an improved performance. Moreover, it may be possible to directly \emph{learn} useful similarity functions by defining the similarity function itself as a neural network trained on a contrastive loss function to minimize differences between variants of the same memory and maximize differences between variants of different ones.

%For a full suite of experiments visualizing the retrieval capacity for all separation function for other performant similarity functions (Manhattan and Euclidean distance) see Appendix F.

%\begin{figure*}[h]
%\centering
%    \begin{subfigure}%{0.33\textwidth}
%        \centering
%        \includegraphics[width=.3\textwid%th]{figures/mnist_N_runs_separation_functions_2.jpg}
        %\caption{MNIST}
%    \end{subfigure}
%        \begin{subfigure}%{0.33\textwidth}
%        \centering
%        \includegraphics[width=.3\textwidth]{figures/N_runs_separation_functions_2.jpg}
        %\caption{Cifar10}
%    \end{subfigure}
%        \begin{subfigure}%{0.33\textwidth}
%        \centering
 %       \includegraphics[width=.3\textwidth]{figures/tiny_N_runs_separation_functions_2.jpg}
        %\caption{Tiny Imagenet}
 %   \end{subfigure}
  %  \caption{The retrieval capacity of the network on retrieving half-masked images using the dot-product similarity function. Plotted are the means and standard deviations of 10 runs. A query was classed as correctly retrieved if the sum of squared pixel differences was less than a threshold of 50. }
%    \label{separation_functions_capacity}
%\end{figure*}

\begin{figure*}[t]
    \centering
    \includegraphics[scale=0.5]{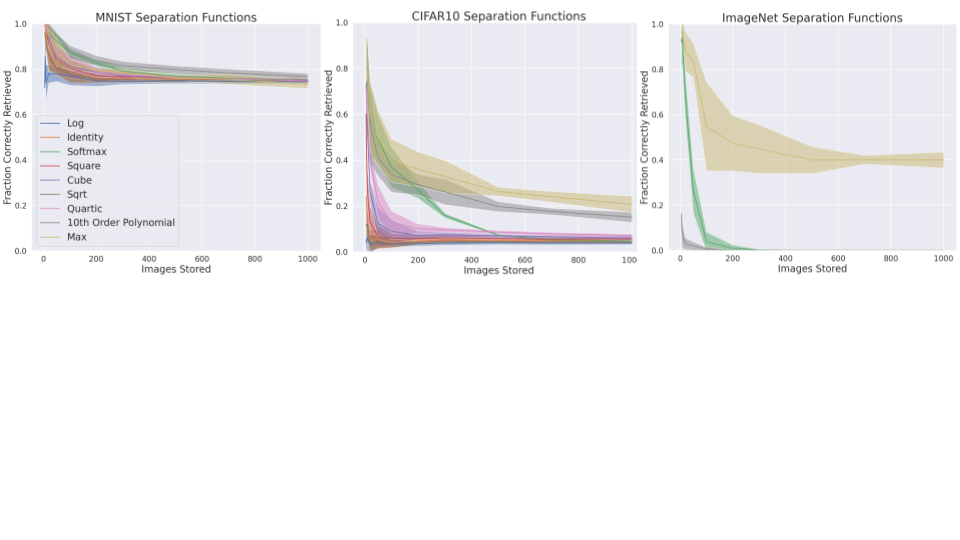}
    \vspace{-5cm}
    \caption{The retrieval capacity of the network on retrieving half-masked images using the dot-product similarity function. Plotted are the means and standard deviations of 10 runs. A query was classed as correctly retrieved if the sum of squared pixel differences was less than a threshold of 50. }
    \label{separation_functions_capacity}
    %\vspace{-0.5cm}
\end{figure*}
%%\vspace{-1cm}

\subsection{Capacity under Different Separation Functions}
In Figure \ref{separation_functions_capacity}, we considered the effect of the separation function on retrieval capacity by measuring the retrieval performance using a fixed similarity function (dot product) for different separation functions (defined in Appendix D). The empirical effect of the separation function on capacity appear to align closely with known theoretical results \citep{demircigil2017model, keeler1988comparison,abu1985information,ma1999asymptotic,wu2012storage}. Namely, that the exponential and max functions have substantially higher capacity than that of other separation functions and that low-order polynomial or lesser separation functions suffer a very rapid decline in retrieval performance as capacity increases. High-order polynomials perform very well, as predicted by the mathematical capacity results in \citep{krotov2016dense, demircigil2017model}. Here, the softmax performs relatively poorly compared to the 10th order polynomial due to the $\beta$ parameter in the softmax being set to $1$, which was done for a fair comparison to other methods. However, as $\beta \rightarrow \infty$, the softmax function tends to the max, so the relative performance of the softmax can be increased by simply increasing $\beta$. The importance of the separation functions, and especially using `high-powered' separation functions such as softmax, max, and a 10th order polynomial increases with the complexity of the data. This is due to the greater level of interference caused by more complex and larger images, which requires a more powerful separation function to numerically push apart the similarity scores.

\subsection{Retrieval under Different Similarity Functions}
\begin{figure*}[t]
    \begin{subfigure}%{0.33\textwidth}
        \centering
        \includegraphics[width=.32\textwidth]{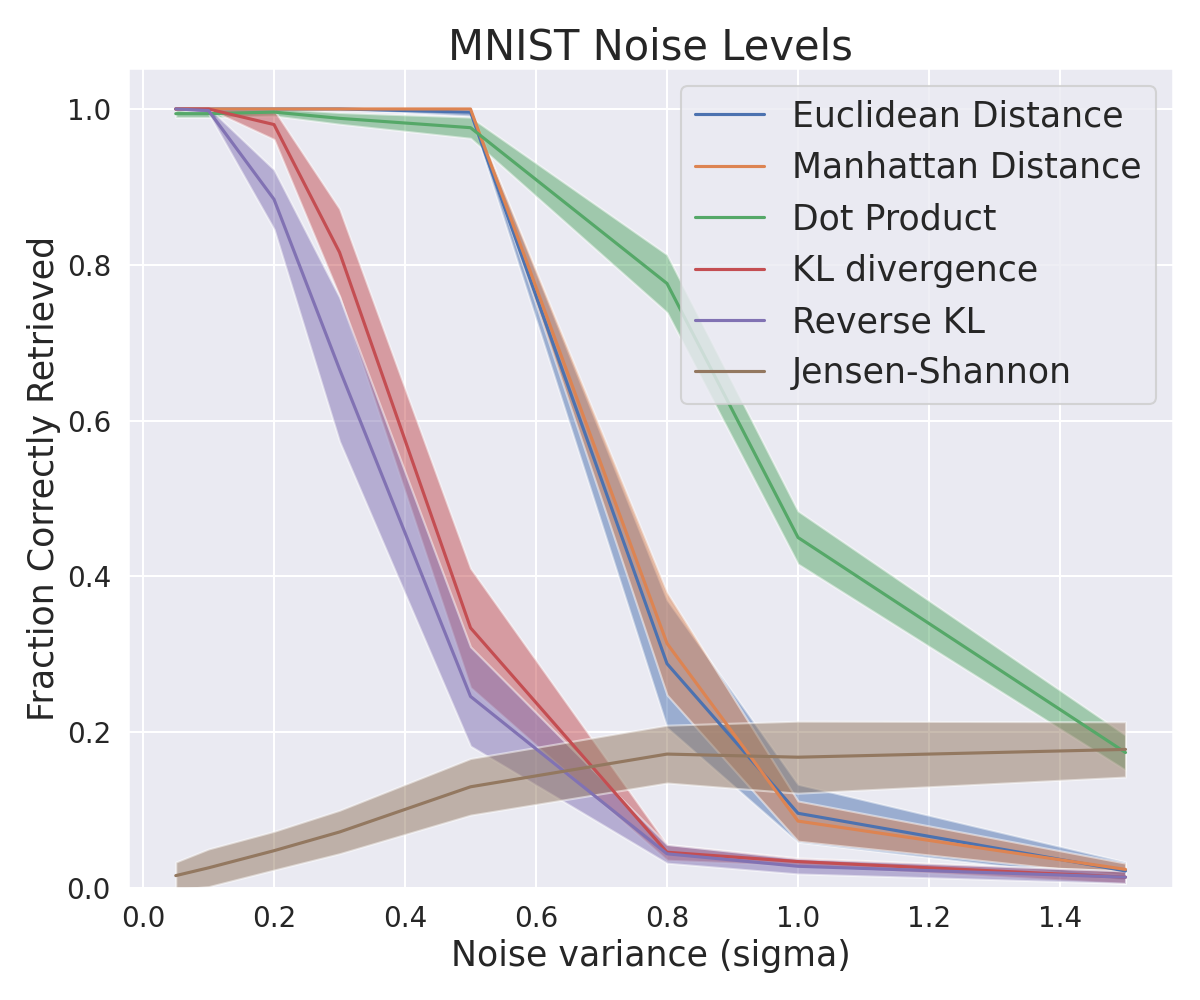}
        %\caption{MNIST}
    \end{subfigure}
        \begin{subfigure}%{0.33\textwidth}
        \centering
        \includegraphics[width=.32\textwidth]{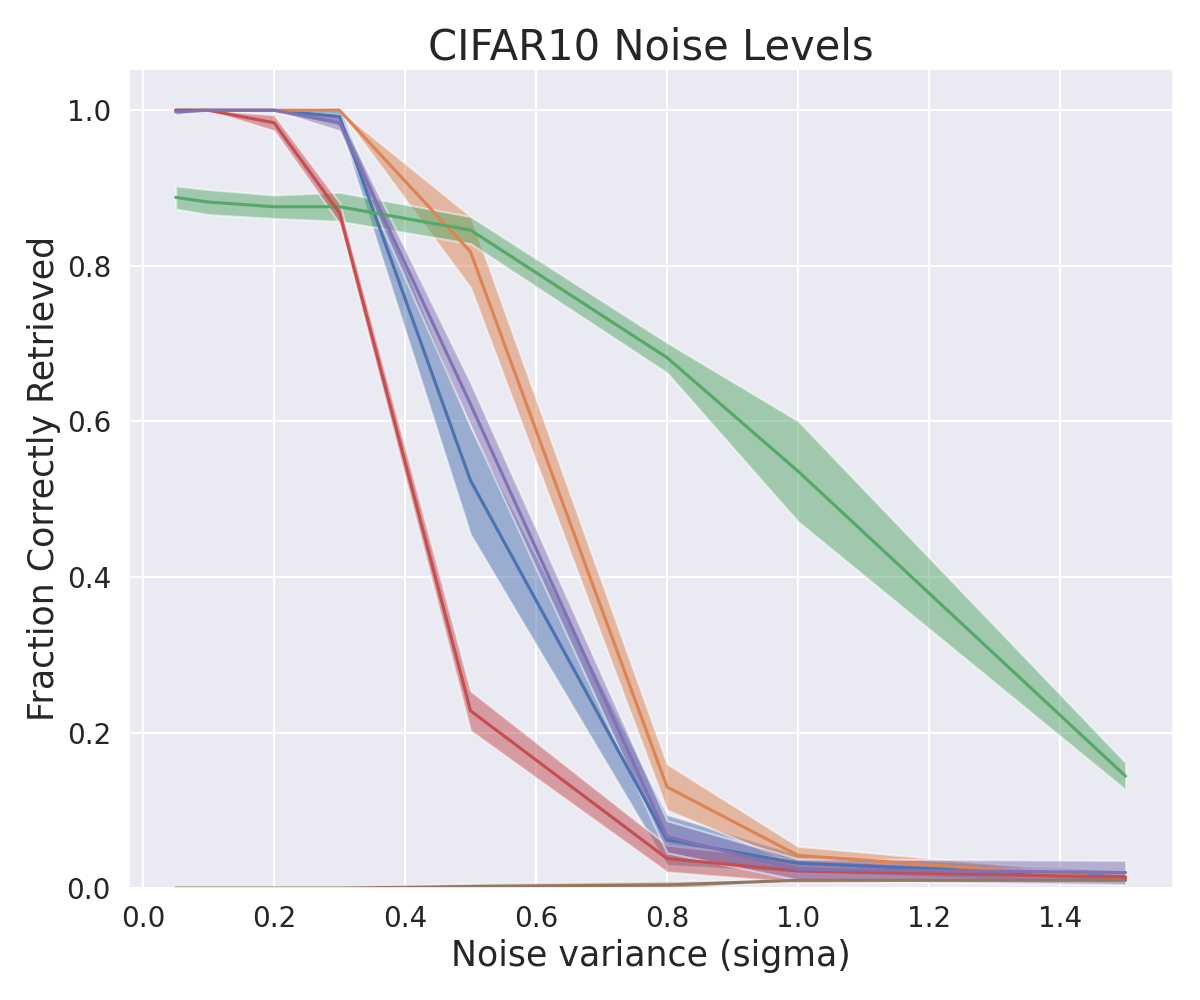}
        %\caption{Cifar10}
    \end{subfigure}
        \begin{subfigure}%{0.33\textwidth}
        \centering
        \includegraphics[width=.32\textwidth]{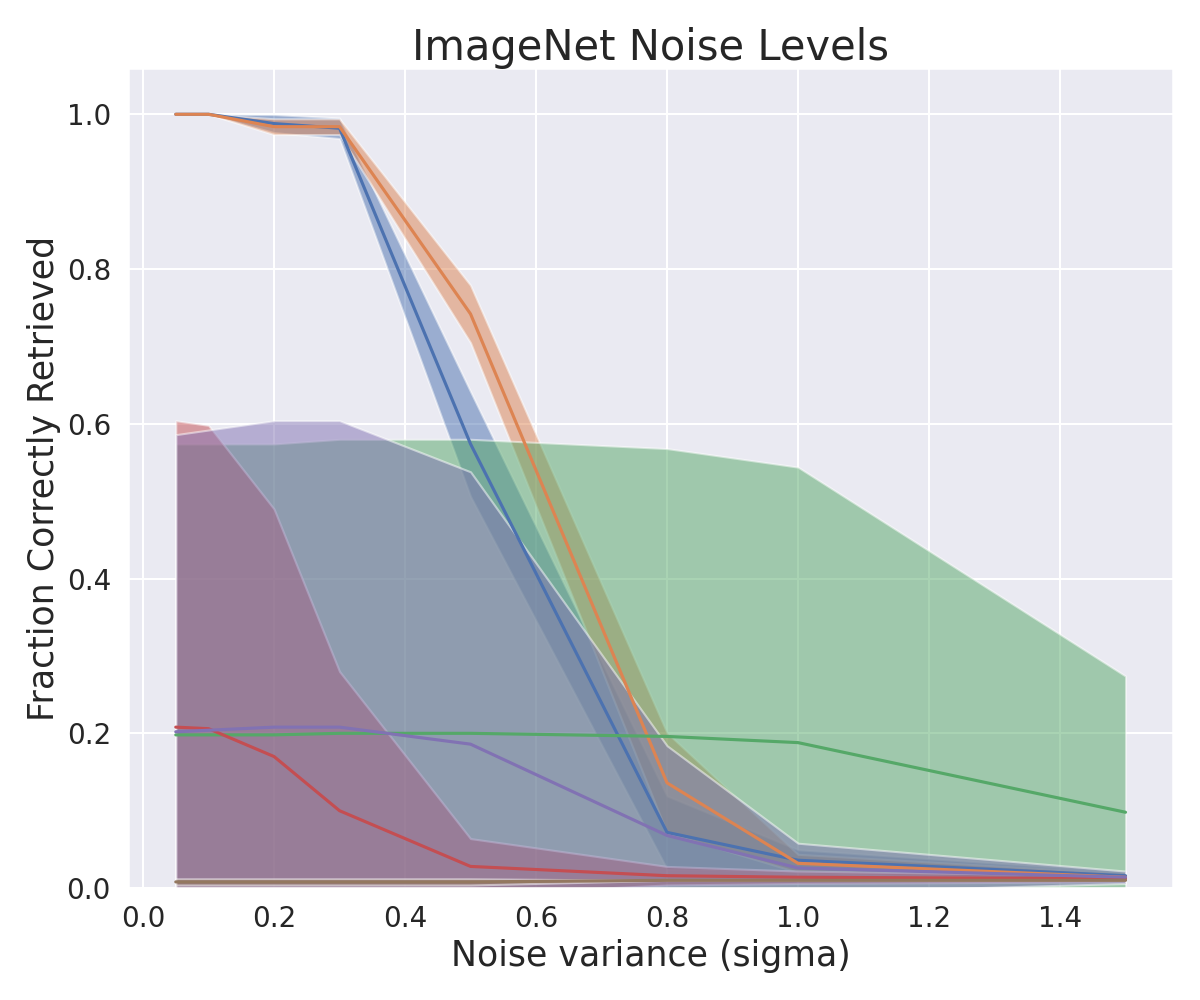}
        %\caption{Tiny Imagenet}
    \end{subfigure}
    \medskip
    \centering
    \begin{subfigure}%{0.33\textwidth}
        \centering
        \includegraphics[width=.32\textwidth]{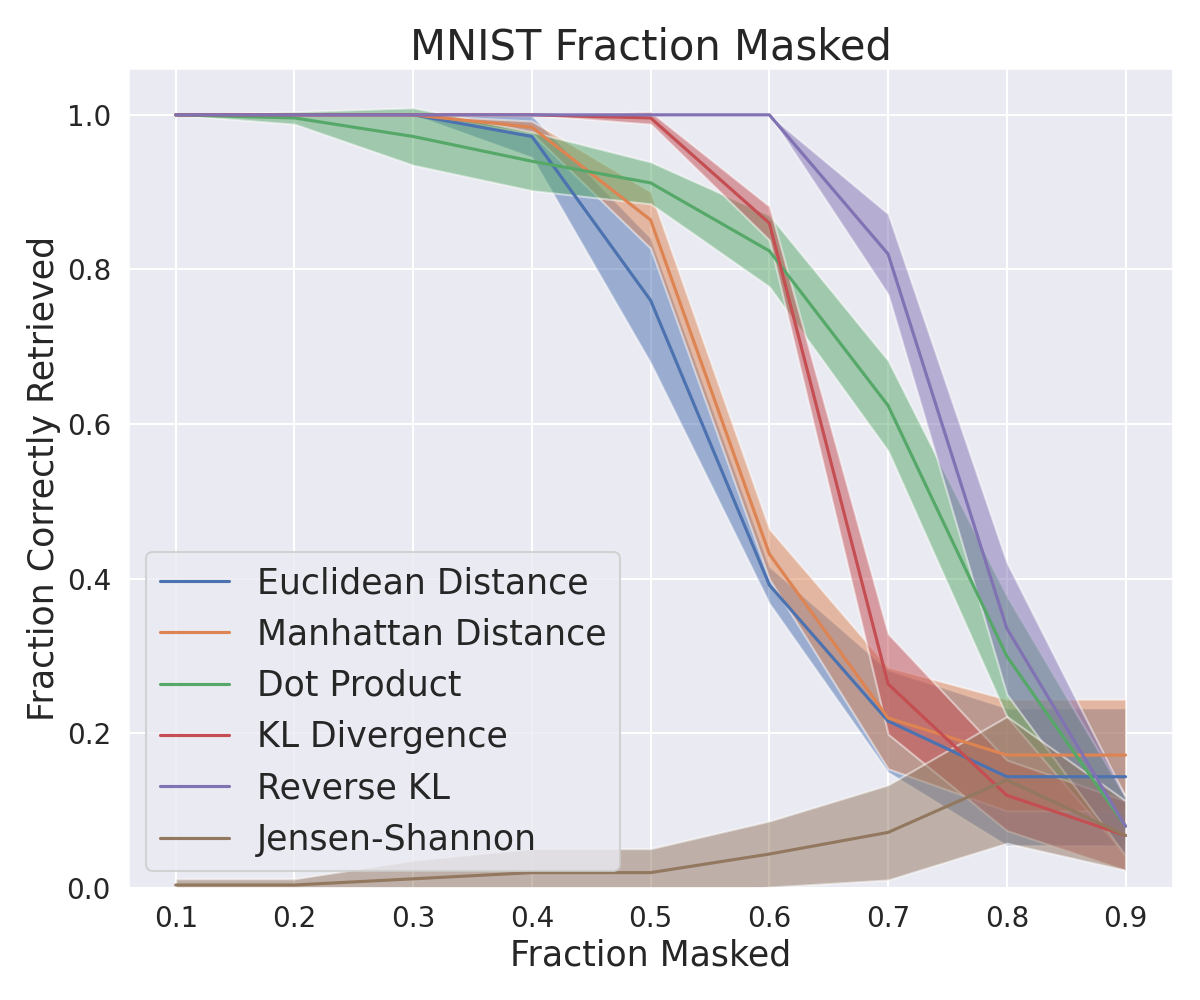}
        %\caption{MNIST}
    \end{subfigure}
        \begin{subfigure}%{0.33\textwidth}
        \centering
        \includegraphics[width=.32\textwidth]{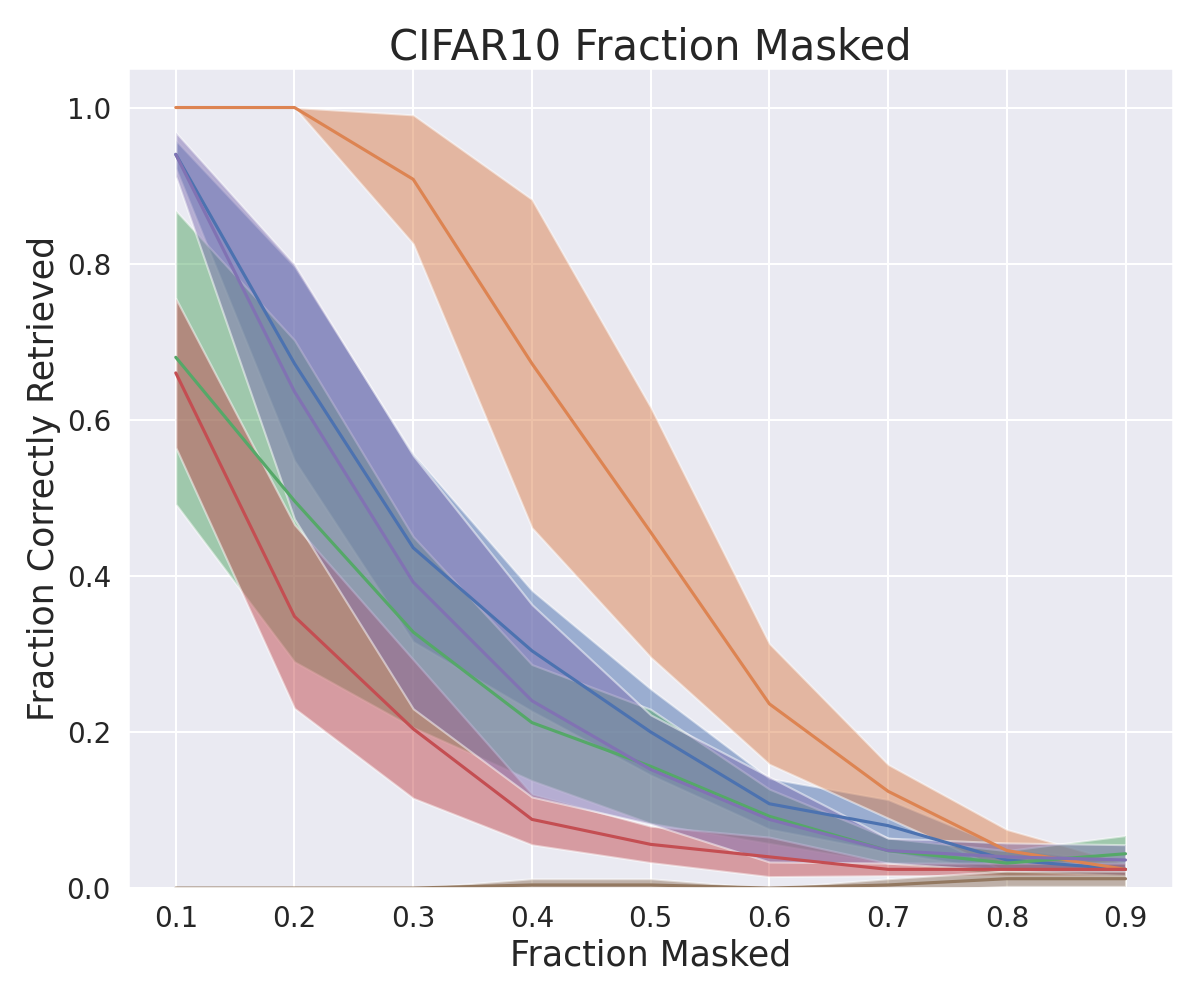}
        %\caption{Cifar10}
    \end{subfigure}
        \begin{subfigure}%{0.33\textwidth}
        \centering
        \includegraphics[width=.32\textwidth]{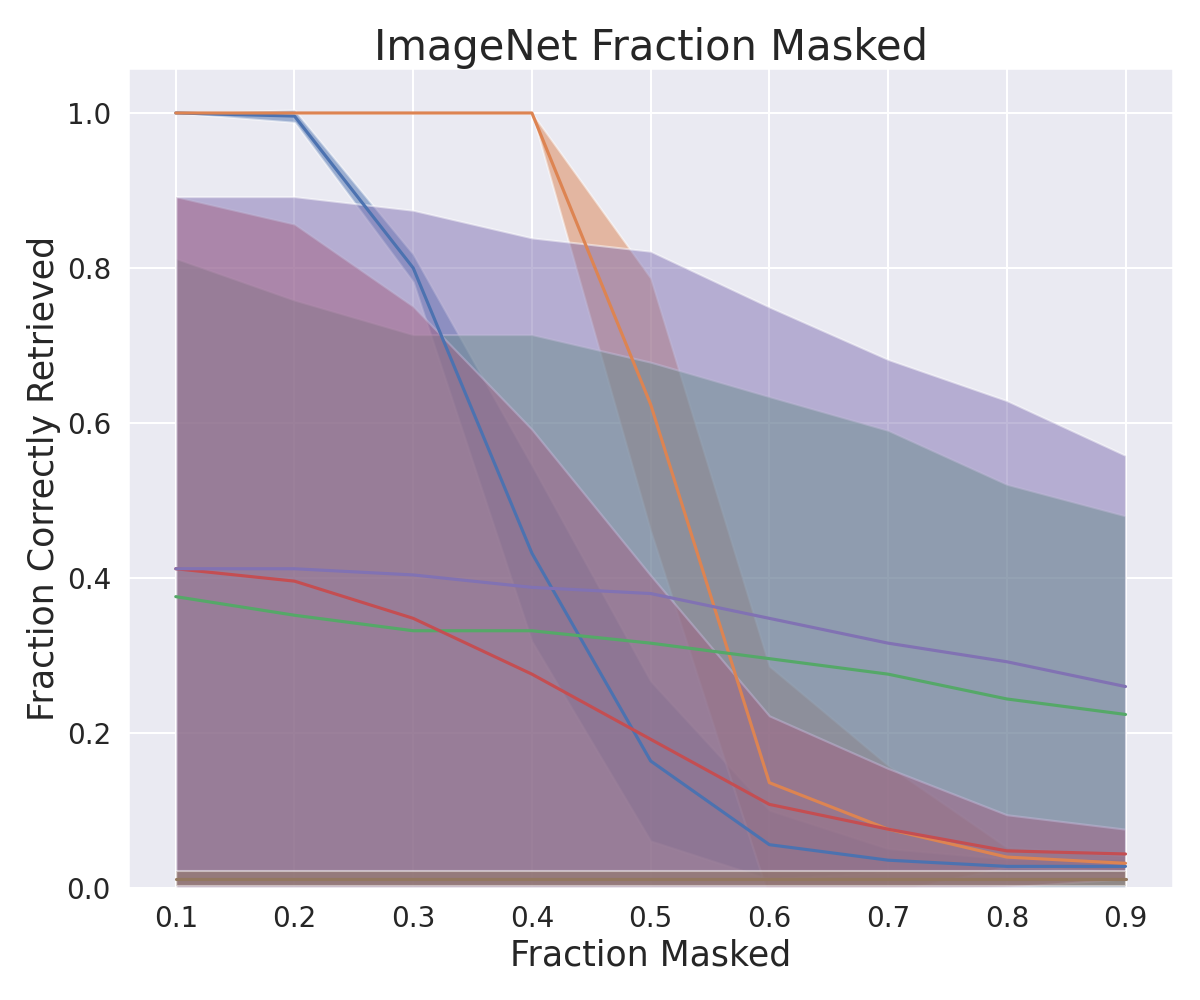}
        %\caption{Tiny Imagenet}
    \end{subfigure}
    \vspace{-0.5cm}
    \caption{\textbf{Top Row}: Retrieval capability against increasing levels of $i.i.d$ added to the query images for different similarity functions. \textbf{Bottom Row}: Retrieval capability against increasing fractions of zero-masking of the query image. The networks used a memory of 100 images with the softmax separation function. Error bars are across 10 separate runs with different sets of memories stored. Datasets used left to right: MNIST, CIFAR, and Tiny ImageNet.}
    \vspace{-0.5cm}
    \label{mask_noise_fig}
\end{figure*}

We also tested (Figure \ref{mask_noise_fig}) the effect of the similarity function on the retrieval capacity of the network for different levels of noise or masking of the query vector, a proxy for the robustness of the memory network. We tested the retrieval capacity on two types of query perturbation: Gaussian noise and masking. In the first case, independent zero-mean Gaussian noise with a specific noise variance $\sigma$ was added elementwise to the query image. As the image pixel values were restricted to lie in the range $[0,1]$, a $\sigma$ of $1$ results in a huge distortion of the original image. With masking, the top $k$ fraction of pixels were set to 0. A fraction of $0.9$ results in only the bottom 10\% of the image being visible in the query vector. Example visualizations different noise levels and masking fractions are given in Appendix C.

We observe in Figure \ref{mask_noise_fig} that the used similarity functions  have strong effects on the robustness of the retrieval under different kinds of perturbations. For independent Gaussian noise, it largely appears that the dot product similarity measures allow for relatively robust reconstructions even up to very high levels of noise, which would make the queries uninterpretable to humans (see Appendix C). The Manhattan distance similarity metric, however, performs better under masking of the image, definitely for relatively small fractions masked, although for Tiny ImageNet, the dot-product similarity function appears to be more robust to extremely high masking fractions. Overall, it appears that the similarity function plays a large role in the degree of robustness of the memory to corrupted queries, but that the same few similarity functions, such as dot product and Manhattan distance, consistently perform well across a range of circumstances.
\vspace{-0.2cm}
\section{Discussion}
\vspace{-0.2cm}
In this paper, we have proposed a simple and intuitive general framework that unifies existing single-shot associative memory models in the literature. Moreover, we have shown that this scheme comes equipped with a set of local neural dynamics and that leads immediately to useful generalizations in terms of varying the similarity function, which led to the discovery of the superior performance of Manhattan distance, which outperforms the state-of-the-art MCHN at various retrieval tasks with complex images. Finally, our general framework lets us see the natural and clear relationship between auto- and heteroassociative memory models, which amounts entirely to the selection of the projection matrix $P$, a fact that has often been unclear in the literature.% While generalizing existing models, our framework also offers a relatively straightforward and simple intuition for why they work, and the computational steps involved in simulating them. 

Our framework gives a clear insight into the two key steps and bottlenecks of current associative memory models. The major bottleneck is the similarity function, which is fundamental to the retrieval capacity and performance of the model, and it is the similarity metric that, assuming a sufficiently powerful separation function, is the main determinant of retrieval performance, especially of noisy or corrupted queries. Here, we only considered single-layer networks, which apply the similarity function directly to raw image data. However, performance may be increased by first feeding the raw queries through a set of preprocessing steps or, alternatively, an encoder network trained to produce a useful latent representation of the input, and then performing associative memory on the latent representations. This naturally leads to a hierarchical scheme for associative memories models, which will be explored in future work. This scheme also has close associations with the field of metric learning \citep{kulis2013metric,yang2007overview}, where we consider the similarity function as defining a metric on the underlying data geometry and then the associative memory simply performs nearest-neighbour matching with this metric. Using predefined similarity functions corresponds to directly defining a metric on the space, however, using a deep neural network to map into a latent space and then performing a similarity scoring on that latent space is equivalent to a learnable metric which is implicitly parametrized by the deep neural network encoder \citep{kaya2019deep}.

A conceptual benefit of our framework is that it makes clear that single-shot associative memory models are simply two-layer MLPs with an unusual activation function (i.e., the separation function), which works best as a softmax or max function, and where the weight matrices directly encode explicit memory vectors instead of being learnt with backpropagation. This leads immediately to the question of whether standard MLPs in machine learning can be interpreted as associative memories instead of hierarchical feature extractors. A crucial requirement for the MLP to function as an associative memory appears to be a high degree of sparsity of intermediate representations (ideally one-hot output) so that an exact memory can be reconstructed instead of a linear combination of multiple memories. With a dense representation at the intermediate layers, no exact memory can be reconstructed, and the network will instead function as a feature detector. This continuum between associative memories and standard MLPs, which depends on the sparsity of the intermediate representations, has resonances in neuroscience, where neural representations are typically highly sparse, as well as to helps contextualize results showing that associative memory capabilities naturally exist in standard machine learning architectures \citep{radhakrishnan2018memorization}.

In terms of the separation function, it is clear that for exact retrieval, the max function is simply the best option, as it removes any interference between different stored memories. The improvement of the separation function is the fundamental cause behind the vast gulf of theoretical capacity and practical performance between the classical HN and the MCHN. It is straightforward to show that with the max separation function, as long as queries are simply uncorrupted copies of the memory images, and that the similarity function of a memory and query has its minimum at the memory (i.e., $\text{sim}(x,x) < \text{sim}(x,y)$ for any $y$), then the $\text{max}$ separation function will achieve a theoretically \emph{infinite} capacity for any fixed size of input query (although, of course, requiring an infinite dimensional memory matrix $M$). However, this theoretical capacity is irrelevant in practice where, for corrupted queries, it is the propensity of the similarity function to detect the right match between query and memory that is the main determinant of retrieval quality.

Our framework also makes a straightforward prediction that the retrieval capacity of hetero- and autoassociative memories are identical for powerful separation functions. This is because the key `work' performed by the memory model is in the first two stages of computing the similarity scores and then separating them, while whether the result is a hetero- or autoassociative memory depends entirely on the projection matrix used to project the resulting separated similarity scores. As such, if the separation scores are nearly a one-hot vector at the correct memory index, the correct image will be `retrieved' by the projection matrix regardless of whether it is a hetero- or autoassociated memory. We verify this prediction by studying the retrieval capacities of  hetero- vs.\ autoassociative MCHNs and HNs in Appendix~B.

Finally, while the capabilities and performance of these associative memory models may seem remote to state-of-the-art machine learning, recent work has begun to link the MHCN with self-attention in transformers \citep{ramsauer2020hopfield}, which has also more recently been linked to SDM \citep{bricken2021attention}. These close links between associative memory models and transformer attention may therefore indicate that improvements in understanding and increasing the effective capacity of such models may also lead to improvements in transformer performance for large-scale machine learning tasks. Perhaps the most interesting avenue lies in testing different similarity functions in transformer models, which, up to now, have almost entirely utilized the dot-product similarity function. This paper, however, has suggested that other similarity functions such as Euclidean and Manhattan distance are also highly competitive with the dot-product similarity and may lead to comparable or superior results when used in transformer self-attention. Preliminary results (Appendix F) suggest that the Manhattan and Euclidean distance similarity functions are competitive with dot product attention in small-scale transformer networks, despite transformer architectures being optimized for the dot product, and suggest that investigating transformer performance more thoroughly with different similarity functions may be an important avenue for future~work.

\section{Code Availability}
Code to reproduce all the experiments and figures reported in this paper is freely available at:  \href{https://github.com/BerenMillidge/Theory\_Associative\_Memory}{https://github. com/BerenMillidge/Theory\_Associative\_Memory}.

\section{Acknowledgements}

We would like to thank Trenton Bricken for many interesting discussions on related topics and Mycah Banks for her help in preparing the figures for this manuscript. Beren Millidge and Rafal Bogacz were supported by the BBSRC grant BB/S006338/1, and Rafal Bogacz was also supported by the MRC grant MC\textunderscore UU\textunderscore 00003/1. Yuhang Song was supported by the China Scholarship Council under the State Scholarship Fund and by a J.P.~Morgan AI Research Fellowship. 
Thomas Lukasiewicz was supported by the Alan Turing Institute under the EPSRC grant EP/N510129/1, the AXA Research Fund, and the EPSRC grant EP/R013667/1.

%\paragraph{\textsc{JPMorgan Chase \& Co.}}
%This research was funded in part by JPMorgan Chase \& Co. Any views or opinions expressed herein are solely those of the authors listed, and may differ from the views and opinions expressed by JPMorgan Chase \& Co. or its affiliates. This material is not a product of the Research Department of J.P. Morgan Securities LLC. This material should not be construed as an individual recommendation for any particular client and is not intended as a recommendation of particular securities, financial instruments or strategies for a particular client. This material does not constitute a solicitation or offer in any jurisdiction.

\bibliography{cites}

%%%%%%%%%%%%%%%%%%%%%%%%%%%%%%%%%%%%%%%%%%%%%%%%%%%%%%%%%%%%%%%%%%%%%%%%%%%%%%%
%%%%%%%%%%%%%%%%%%%%%%%%%%%%%%%%%%%%%%%%%%%%%%%%%%%%%%%%%%%%%%%%%%%%%%%%%%%%%%%
% APPENDIX
%%%%%%%%%%%%%%%%%%%%%%%%%%%%%%%%%%%%%%%%%%%%%%%%%%%%%%%%%%%%%%%%%%%%%%%%%%%%%%%
%%%%%%%%%%%%%%%%%%%%%%%%%%%%%%%%%%%%%%%%%%%%%%%%%%%%%%%%%%%%%%%%%%%%%%%%%%%%%%%
\newpage
\appendix
\onecolumn

\section*{Appendix A: Proof of Energy Function Being a Lyapunov Function of the Dynamics}

In this appendix, we demonstrate that the energy function is a Lyapunov function of the dynamics. This means that by running the dynamics forward in time, the value of energy function is guaranteed to decrease. To do so, we simply compute the time derivative of the energy function and show that it must be negative:
\begin{align*}
    \frac{dE}{dt} &= \sum_i  \sum_j v_i \frac{\partial \mathcal{L}^2}{\partial v_i \partial v_j} \frac{dv_i}{dt} + \frac{dv_i}{dt}\frac{\partial \mathcal{L}_v}{\partial v_i} - \frac{\partial \mathcal{L}_v}{\partial v_i}\frac{dv_i}{dt} + \sum_i \sum_j h_i \frac{\partial \mathcal{L}_h^2}{\partial h_i \partial h_j} \frac{dh_i}{dt} + \\ &\sum_i \frac{\partial \mathcal{L}_h}{\partial h_i} \frac{dh_i}{dt} - \sum_i \frac{dh_i}{dt} \frac{\partial \mathcal{L}_h}{\partial h_i} - \sum_i \sum_j \frac{\partial \mathcal{L}^2_v}{\partial v_i \partial v_j} \frac{\partial \text{sim}(M_{i,j}, v_i)}{\partial v_i} \frac{dv_i}{dt} - \sum_i \sum_j \text{sim}(M_{i,j}v_i) \frac{\partial \mathcal{L}^2_h}{\partial h_i \partial h_j} \frac{dh_i}{dt} \\
    &= \sum_i \sum_j v_i \frac{\partial \mathcal{L}^2_v}{\partial v_i \partial v_j} \frac{dv_i}{dt} + h_i \frac{\partial \mathcal{L}^2_h}{\partial h_i \partial h_j} \frac{dh_i}{dt} +  \frac{\partial \mathcal{L}^2_v}{\partial v_i \partial v_j} \frac{\partial \text{sim}(M_{i,j}v_i)}{\partial v_i} \frac{dv_i}{dt} - \text{sim}(M_{i,j},v_i) \frac{\partial \mathcal{L}^2_h}{\partial h_i \partial h_j} \frac{dh_i}{dt} \\
    &= \sum_i \sum_j \frac{dv_i}{dt}\frac{\partial \mathcal{L}^2_v}{\partial v_i \partial v_j}\big[v_i - \frac{\partial \text{sim}(M_{i,j},v_i)}{\partial v_i} \big] + \frac{dh_i}{dt}\frac{\partial \mathcal{L}^2_h}{\partial h_i \partial h_j} \big[ h_i - \text{sim}(M_{i,j},v_i) \big] \\
    &= - \sum_i \sum_j \frac{dv_i}{dt} \frac{\partial \mathcal{L}^2_v}{\partial v_i \partial v_j} \frac{dv_i}{dt} + \frac{dh_i}{dt} \frac{\partial \mathcal{L}^2_h}{\partial h_i \partial h_j} \frac{dh_i}{dt}\,, \numberthis
\end{align*}
which is clearly always negative as long as the Hessians of the activation functions are positive definite. In the usual case of elementwise activation functions, this requires that the functions be monotonically increasing. Note that in this derivation, we have assumed that the input currents are constant $\frac{dI}{dt} =0$, the fact that the derivative of the Langrangians can be defined by the chain rule as $\frac{d \mathcal{L_v}}{dt} = \frac{\partial \mathcal{L}_v}{\partial v}\frac{dv}{dt}$, and the definition of the dynamics of the visible and hidden neurons.

\newpage

\section*{Appendix B: Heteroassociative Hopfield Networks}

In this appendix, we follow up on the discussion in Section 1.1 and demonstrate that both the MCHN and the HN can be straightforwardly extended to perform heteroassociative memory retrieval with no impact on performance compared to autoassociativity. This is done simply by replacing the memory matrix in the projection step by a different memory matrix which represents the other memories that must be heteroassociated with the main memories. If we define the memory matrix as $M$ and the heteroassociative projection matrix as $P$, this results in the following update rule for the MCHN:
\begin{align}
    z^* = P \sigma(M \cdot q)
\end{align}
and the following update rule for the HN:
\begin{align}
    z^* = \text{sign}(\tilde{T} \cdot q)\,,
\end{align}
where the heteroassociative memory matrix $\tilde{T}$ can be written as $PM^T$. The reason for the negligible performance difference between auto- and heteroassociation is that all the `difficult' computation that can lead to mis-reconstructions occurs during the computation of the similarity scores and the application of the separation function. Once the set of similarity scores is computed, these scores simply select the linear combination of rows of the projection matrix that is to be reconstructed. Whether this projection matrix is the same as the memory matrix $M$, or some other autoassociation matrix $P$ is immaterial.

\begin{figure}[H]
    \centering
    \includegraphics[width=\textwidth]{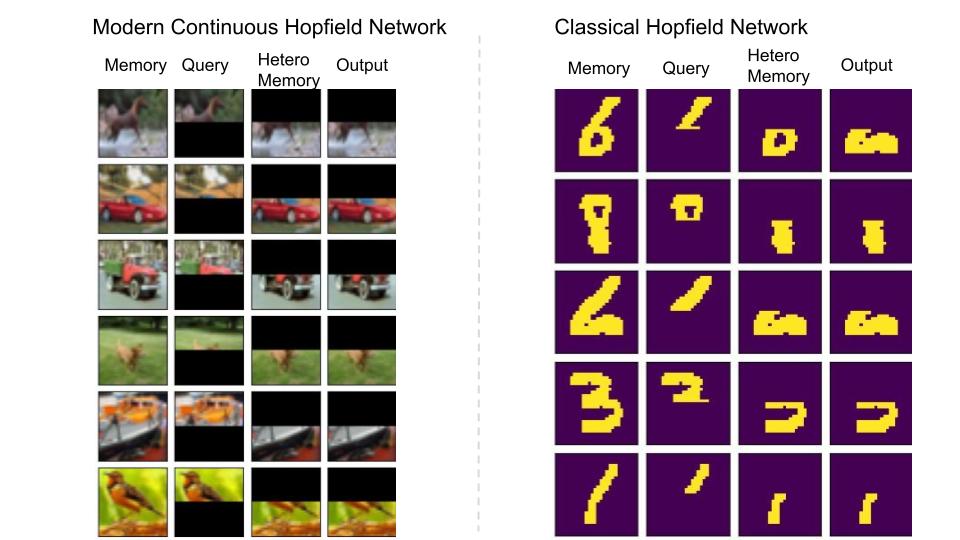}
    \caption{Examples of heteroassociative memory retrieval for both the MCHN and the HN. The networks were queried with one half of either a CIFAR10 image for the MCHN or a binarized MNIST digit for the HN. The autoassociated memory was the other half of the original memory image. On the left, the MCHN achieves perfect heteroassociation, even of challenging CIFAR10 images, due to its superior capacity to the HN. On the right, the HN achieves substantially correct heteroassociations but occasionally misreconstructs an~image. }
    \label{heteroassociation_example_fig}
\end{figure}

An additional direct consequence of our theoretical framework is that there should be effectively no difference in hetero- vs.\ autoassociative memory retrieval performance for any feedforward memory model, since all the `work' is already done in computing the similarity and separation functions, while the difference between auto- and heteroassociative memories occurs only at the projection matrix. We verify this prediction here by comparing the retrieval capacity of auto- and heteroassociative MCHNs and HNs as the memory size increases and find them to be statistically identical.

%\begin{figure*}[t]
%\centering
%    \begin{subfigure}%{0.5\textwidth}
%        \centering
%        \includegraphics[width=.45\textwidth]{figures/mnist_classical_hopfield_auto_hetero_comparison_gaussian.jpg}
        %\caption{Classical Hopfield}
%    \end{subfigure}
%        \begin{subfigure}%{0.5\textwidth}
%        \centering
%        \includegraphics[width=.45\textwidth]{figures/cifar_10_auto_hetero_comparison_gaussian.jpg}
        %\caption{MCHN}
%    \end{subfigure}
%    \caption{Comparison of autoassociative vs Heteroassociative MCHN and classical hopfield networks on retrieval task. For both, given a corrupted image, the heteroassociative task was to retrieve only the bottom half. The MCHN was queried with Cifar10 images corrupted with Gaussian noise of variance $0.5$. The classical hopfield network was tested on binarized mnist images where the query was the top half of the image. Error bars are the standard deviations of the retrieval capacity over 10 runs. The performance of the classical hopfield network is extremely poor due to interference between memories caused by its identity separation function. In both cases, the differences between autoassociative and heteroassociative capacity are negligible}
%    \label{hetero_auto_capacity_graph}
%\end{figure*}

\begin{figure}[H]
    \centering
    \includegraphics[scale=0.5]{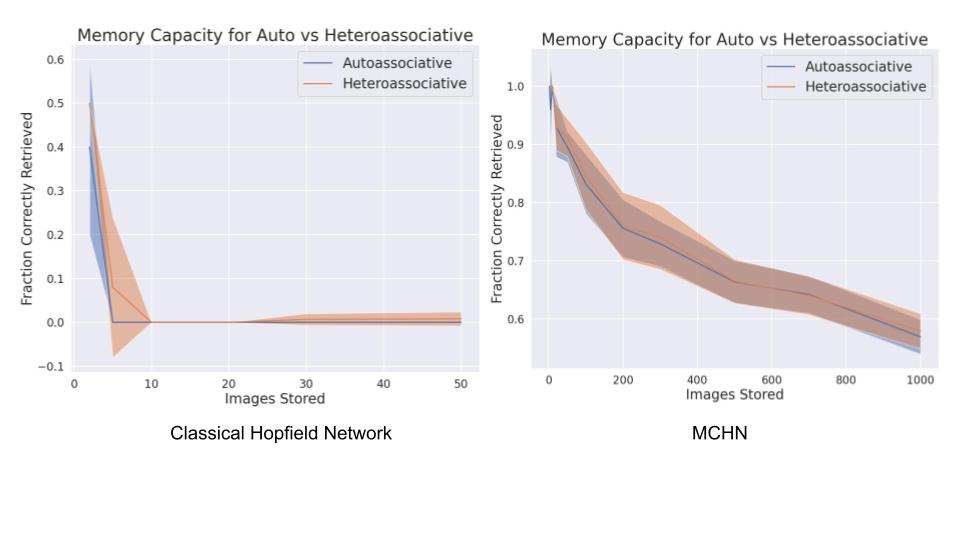}
    %\vspace{-2cm}
    \caption{Comparison of auto- vs.\ heteroassociative MCHN and HNs on retrieval task. For both, given a corrupted image, the heteroassociative task was to retrieve only the bottom half. The MCHN was queried with CIFAR10 images corrupted with Gaussian noise of variance $0.5$. The HN was tested on binarized MNIST images where the query was the top half of the image. Error bars are the standard deviations of the retrieval capacity over 10 runs. The performance of the HN is extremely poor due to interference between memories caused by its identity separation function. In both cases, the differences between auto- and heteroassociative capacity are negligible}
    \label{hetero_auto_capacity_graph}
\end{figure}

\newpage

\section*{Appendix C: Reconstructions under Varying Levels of Image Perturbation}

In this appendix, we include example reconstructions under varying levels of query degradation as an example of the two kinds of query degradation examined in the main text. These are addition of random Gaussian noise with varying variance, and masking (with zeros) of a given fraction of the query image. We present sample reconstructions of an associative memory network (with Manhattan distance similarity and softmax separation functions) under both different levels of noise variances and fractions of the image masked out. The values shown here are the same as in the capacity robustness plots (Fig.~\ref{capacity_graph_figure}), so that an intuitive picture of the difficulty of the network's tasks can be gauged.

\begin{figure}[H]
    \centering
    \includegraphics[width=\textwidth]{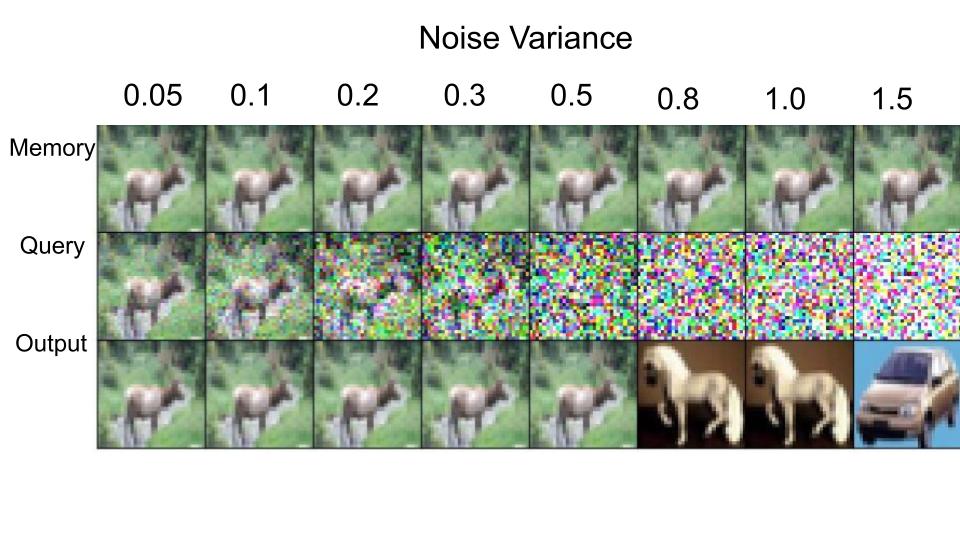}
    \caption{Examples of reconstruction of an associative memory network using the Manhattan distance similarity function and the softmax separation function. The network achieves a perfect performance up to a noise variance of about $0.5$, which visually is an extremely distorted version of the original image. For high variances, the reconstructions are incorrect, however, a feature of the MCHN is that the attractors span the space such that any input pattern, even a meaningless one, will eventually be classed as a given pattern.}
    \label{reconstructions_noise_variance}
\end{figure}

\begin{figure}[H]
    \centering
    \includegraphics[width=\textwidth]{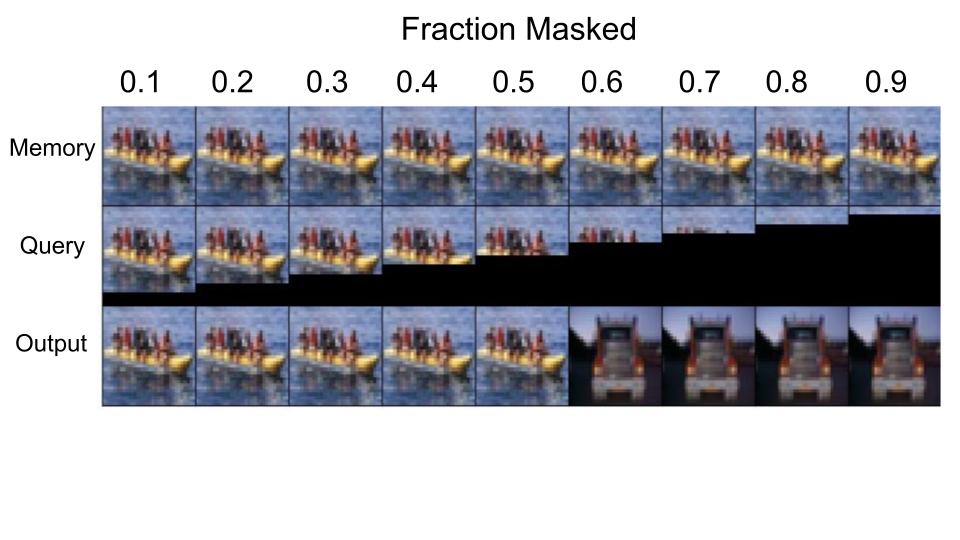}
    \caption{Examples of reconstruction of an associative memory network using the Manhattan distance similarity function and softmax separation function. The network achieves a perfect performance up to a fraction masked of $0.5$, while reconstructions afterwards are incorrect. Interestingly, visually to a human this task is much easier than the Gaussian noise distortion, but the network finds denoising the Gaussian noise significantly easier. This may be due to the design of the similarity functions for which the noisy images are `closer' in space to the memory than images with half or more of the image as zeros, which of course generates large errors for all the zero pixels.}
    \label{reconstructions_masked}
\end{figure}

\newpage

\section*{Appendix D: Suite of Similarity and Separation Functions}
The list of similarity functions tested is presented in Table \ref{similarity_function_table} below.
\newline 

\begin{table}[H]
\centering
\caption{Similarity Functions}

\smallskip 
\begin{tabular}{ p{5cm}p{8cm} }
 \toprule
 \textbf{Similarity Function} & \textbf{Definition} \\
 \midrule
 Euclidean Distance   & $\text{sim}(M,q) = \sum (M - q)^2$  \\
 Manhattan Distance &   $\text{sim}(M,q) = \sum \text{abs}(M-q)$ \\
 Dot Product & $\text{sim}(M,q) = M.q$ \\
 KL Divergence & $\text{sim}(M,q) = \sum q \ln \frac{M}{q}$  \\
 Reverse KL Divergence & $\text{sim}(M,q) = \sum M \ln \frac{M}{q}$ \\
 Jensen-Shannon Divergence & $\text{sim}(M,q) = \frac{1}{2}KL[M||q] + \frac{1}{2}KL[q||M]$ \\
 \bottomrule
\end{tabular}
\label{similarity_function_table}
\end{table}

Similarly, the list of separation functions tested is given in the Table \ref{separation_function_table} below.
\newline

%\begin{center}
\begin{table}[H]
\centering
\caption{Separation Functions}

\smallskip 
\begin{tabular}{ p{5cm}p{8cm} }
 \toprule
  \textbf{Separation Function} & \textbf{Definition} \\
 \midrule
 Identity  & $\text{sep}(x) = x$  \\
 Square &   $\text{sep}(x) = x^2$ \\
 N-th Order Polynomial &$\text{sep}(x,n) = x^n$ \\
 Log & $\text{sep}(x) = \ln x$  \\
 Softmax &$\text{sep}(x, \beta) = \frac{e^{-\beta * x}}{\sum e^{-\beta * x}}$ \\
 Max & $\text{sep}(x) = \text{max}(x)$ \\
 \bottomrule
\end{tabular}
\label{separation_function_table}
\end{table}
%\end{center}

\newpage

\section*{Appendix E: Numerical and Simulation Notes}
Two key issues for making a fair comparison between similarity functions is the numerical effects of scaling and the direction of similarity. Firstly, implemented naively, the similarity metrics often have different characteristic scales, i.e., one measure might naturally return values that are much larger or smaller than another. This would then change the effect of the separation function and thus the reconstruction accuracy. For instance, a method that returned high similarity values would often be easier to separate than one which returned small ones. To address this problem, we normalized the similarity scores of each similarity function so that they would sum to 1. A second problem is the direction of similarity, namely,  whether the similarity function returned high or low values for similar inputs. Similarity measures such as the dot product give large dot products for similar inputs, while distance measures such as the Euclidean distance give small distances for similar inputs. To address this, for distance metrics, we instead returned the normalized reciprocal of the distances, so that large inverse distances correspond to a high similarity. Thus, all similarity functions returned their scores in a standardized normalized format whereby larger scores represented larger degrees of similarity, and the sum of all scores was normalized to one. The outputs of the separation function are also normalized such that their sum is 1.

To compute the number of correct retrievals, given a memory matrix and a query vector, we first implemented the reconstruction through the main equation $z = M^T \text{sep}(\text{sim}(M, q))$, where $sim$ is defined to output scores in the normalized format. The input vector $q$ is derived by simply flattening the input image. The memory matrix $M$ consists of a series of flattened vectors for the stored images $M = [m_1, m_2, \dots ]$, where $m_1$ is the flattened vector of a memory image. Once the reconstruction $z$ was computed, we compared it to the original image and computed a reconstruction score based on the sum of the squares of the elementwise differences between the reconstruction and the true image $L = \sum_i (z_i - z^*_i)^2$. If the sum of squares was less than a threshold (here, we used 50), then the image was classed as being correctly retrieved and otherwise not. The threshold value was chosen empirically to allow reconstructions that are almost indistinguishable by eye from the true input, while ruling out incorrect and poor reconstructions.

\subsection*{E.1. Dot-Product Similarity}

A key weakness of the dot-product similarity metric is that it is not invariant to the norm of the vectors. This means that the similarity computed depends heavily on the vector norms often more so than the similarities. Mathematically, this results in the fact that it is not the case that $x^Tx \geq x^Tz$, where $z$ is any other vector. What this means is that two completely different inputs can have a higher dot product similarity than the input dotted with itself. This does not happen with other similarity metrics such as the Euclidean distance where the minimum distance of $0$ is achieved when the query vector and a memory vector are identical. This occurs because the $z$ vector may have a larger norm than the $x$ vector. This problem in practice leads to catastrophically poor performance of the dot-product similarity metric, especially on dense color images like  the CIFAR and Tiny ImageNet datasets, where the un-normalized dot product simply tends to simply measure the degree of high pixel values in an image. To alleviate this issue,  results in the paper are instead reported using a normalized dot-product similarity function defined as
\begin{align}
    \text{dot}(X,z) = \text{norm}(\text{norm}(X) \cdot \text{norm}(z))\,,
\end{align}
where $\text{norm}(x) = \frac{x}{\sum x}$ simply normalizes the entries in the vector to sum to 1,  and where the norm on the memory matrix $X$ is taken for each row (i.e., stored vector) independently. The dot product of the two normalized vectors is then normalized again for numerical reasons, since otherwise the similarity scores computed were often extremely small leading to numerical issues and poor performance with the separation function. 

A similar normalization was also performed for the similarity functions that involved a KL divergence, which possesses a probabilistic interpretation, and thus all the input vectors were normalized, so as to sum to one, and thus preserve an interpretation in terms of probability distributions.

\newpage

\section*{Appendix F: Transformer Experiments}

To test whether the insights gained from this framework might apply to large-scale machine learning in the form of improving transformer attention, we implemented transformer layers using various similarity functions. Mathematically, we modified the transformer update rule to
\begin{align}
    z = V \sigma(\beta \cdot \text{sim}(K,Q)),
\end{align}
where $V$, $K$, and $Q$ are the Value, Key, and Query matrices of transformer attention, $\beta$ is the softmax temperature, $\sigma$ is the softmax function, and $\text{sim}$ is the similarity function. All other aspects of the transformer architecture remained the same.

We utilized an encoder-decoder transformer architecture with 2 residual blocks. Each residual block included a modified attention layer, an MLP layer, and batch normalization. The transformers were trained on the Wikitext dataset using the ADAM optimizer. The MLP layer had a hidden dimension of 200, and the embedding dimension was also 200. Two attention heads were used. A batch size of 20 was used.

\begin{figure}[H]
    \centering
    \includegraphics[scale=0.3]{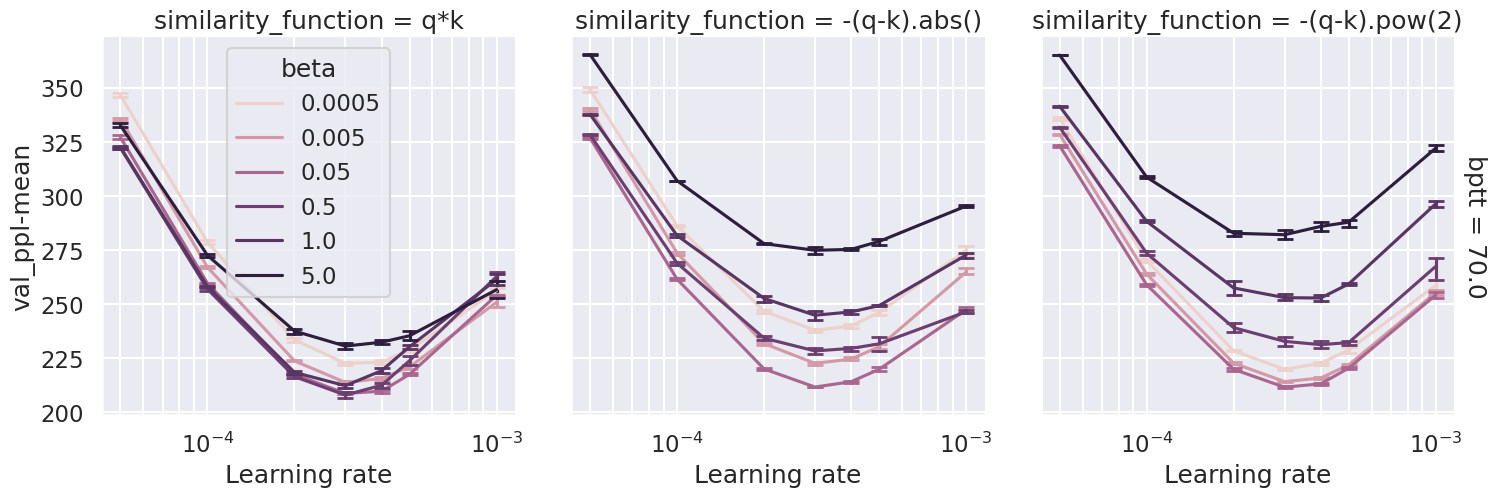}
    \caption{Achieved perplexity on the WikiText dataset using transformer models with varying similarity functions across a range of learning rates. All successful similarity functions achieved similar results although the absolute value and Euclidean distance similarity functions appeared more sensitive to choices of the $\beta$ hyperparameter. }
    \label{transformer_experiments}
\end{figure}

Although small-scale, from the preliminary results, it appears that the similarity function used actually makes relatively little difference for the final learning performance of the transformer. This may suggest that despite recent works interpreting attention through the lens of heteroassociative memories \citep{ramsauer2020hopfield,bricken2021attention}, transformers are not particularly functioning as associative memories in that the learnt $K$ and $Q$ matrices do not directly correspond to learnt `memories' but rather that  the success of the transformer architecture is simply due to the mathematical structure of the update rule --- for instance, the multiplicative interactions, which have also been shown to be crucial in non-transformer contexts, such as in the MLP-mixer architecture \citep{tolstikhin2021mlp,jayakumar2020multiplicative}. Supporting evidence for this comes from the fact that the post-softmax attention outputs are typically dense and the softmax temperature is often relatively high, meaning that the system is not reconstructing exact memories as the MCHN does. Sparsifying self-attention outputs (for instance, with a large inverse temperature or with an explicit \emph{max} or \emph{top-k} function  \citep{gupta2021memory} replacing the softmax), however, will have the effect of making the self-attention function much more similar to an associative memory that retrieves specific memories given a cue.

Additionally, it could simply be that given that the $K$, $Q$, and $V$ matrices are all learnable, that backpropagation can simply route around the different similarity functions, and specialize representations for a given similarity function. If this is the case, then it may indicate that substantial computational savings may be obtained by devising similarity functions that are less expensive than the dot product currently used. Much work in this area is already underway and many computationally cheaper simplifications of the transformer have been proposed \citep{kitaev2020reformer,wang2020linformer,tay2020efficient}.

\newpage

\section*{Appendix G: Effect of Error Threshold}

\begin{figure}[H]
    \centering
    \includegraphics[scale=0.4]{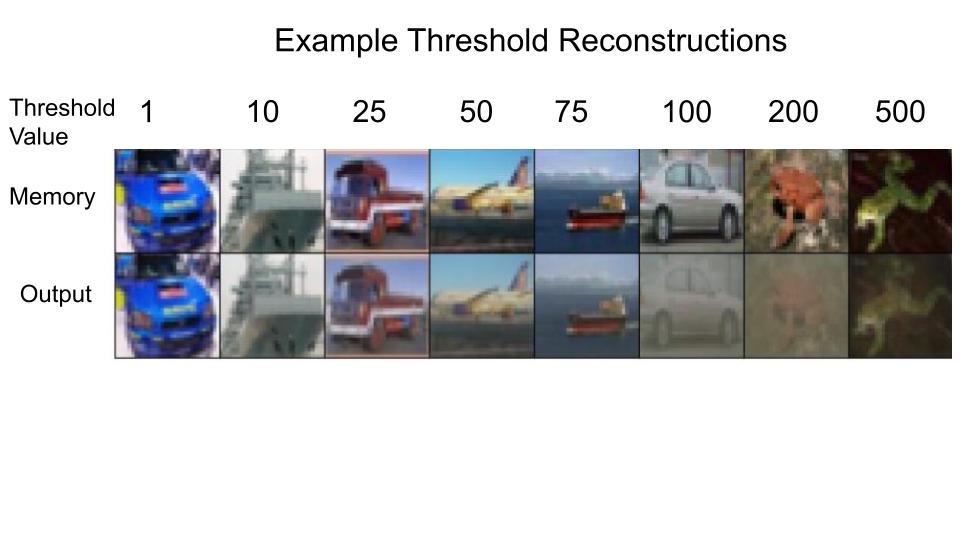}
    \caption{Examples of reconstructed images by a Manhatten distance associative memory model according to various threshold values. Note that the correct retrieval is typically achieved with significantly higher threshold values, but the relatively low threshold enforces \emph{exact} as opposed to blurry reconstructions.}
    \label{threshold_examples}
\end{figure}

For the simulations in this paper, the error threshold was set somewhat arbitrarily as the sum of squared pixel values between the reconstruction and the original image, needing to be less than $50$ for the reconstruction to be classified correctly. Although heuristic, this threshold roughly corresponds to retrieving the correct image and tolerating some fuzziness in reconstruction, but not too much. As an example of reconstructions with different squared differences, see Figure \ref{threshold_examples}.

To validate that the exact value of the threshold had relatively little affect on our results, we demonstrate the retrieval accuracy on CIFAR10 of the three similarity functions over a range of potential thresholds, noting that the curve is relatively flat at the threshold of $50$ chosen and that all similarity functions are affected approximately equally by the change in the threshold --- unsurprisingly, with more images being  correctly retrieved with a larger threshold value. 

\begin{figure}[H]
    \centering
    \includegraphics[scale=0.35]{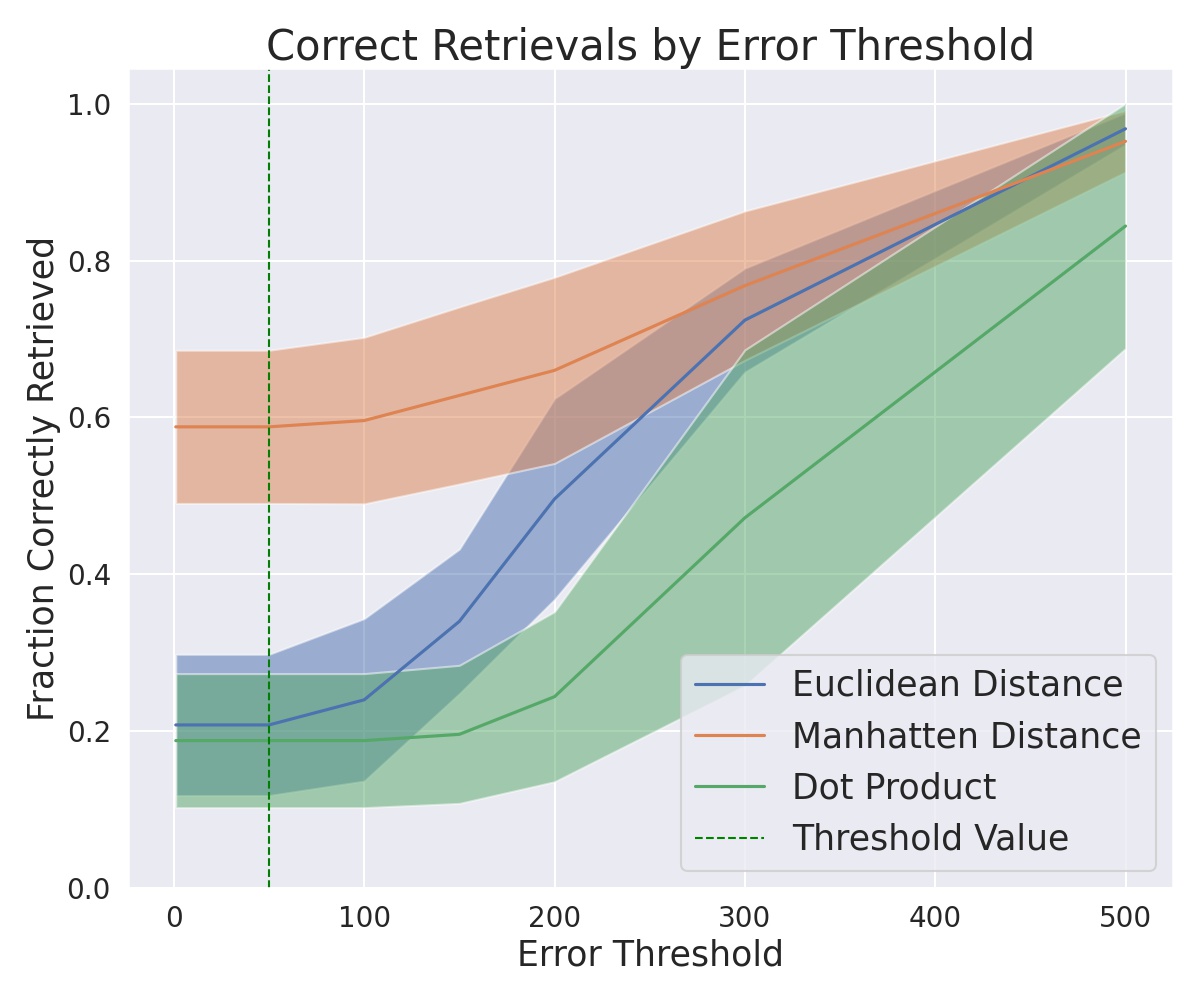}
    \caption{Percentage of correct retrieval against error threshold. Note that the threshold chosen lies at the flat part of the graph, although unsurprisingly, correct retrievals increase rapidly with large thresholds.}
    \label{error_threshold_sweep}
\end{figure}

\newpage

\section{Appendix H: Additional Image Perturbations}

\begin{figure}[H]
    \centering
    \includegraphics[scale=0.4]{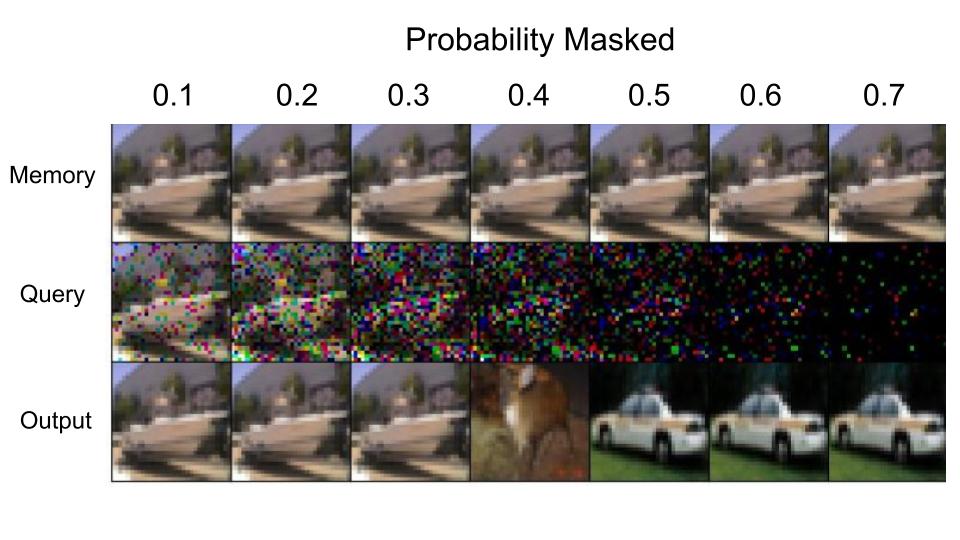}
    \caption{Example queries and reconstructions for the random masking perturbation.}
    \label{random_masking_perturbation}
\end{figure}

\begin{figure}[H]
    \centering
    \includegraphics[scale=0.4]{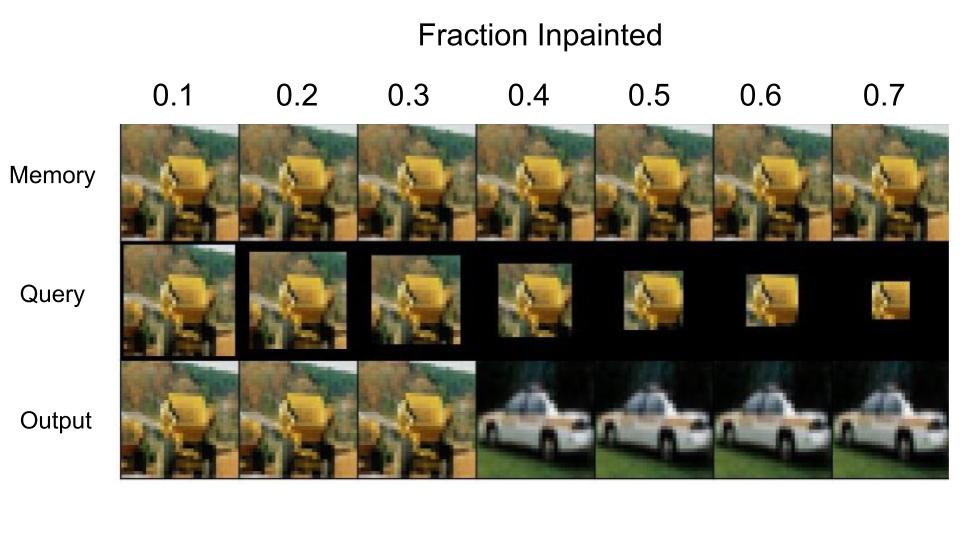}
    \caption{Example queries and reconstructions for the image inpainting perturbation.}
    \label{inpaining_perturbation}
\end{figure}

To test whether the same effects of similarity function are maintained on a wider set of image perturbations, in this appendix, we additionally present an analysis of two further perturbations --- \emph{random masking}, where randomly chosen pixels on the image are masked out (set to zero) with a certain probability, and inpainting, where the edges of the image are masked out, and the edges must be reconstructed from the center. Examples of both perturbations for differing levels of pixel masking are presented in Figures \ref{random_masking_perturbation} and \ref{inpaining_perturbation}, respectively.

We performed equivalent capacity tests of the three main similarity functions proposed (Manhattan and Euclidean distance and dot-product similarity) and plotted the results below,

\begin{figure}[H]
    \centering
    \includegraphics[scale=0.5]{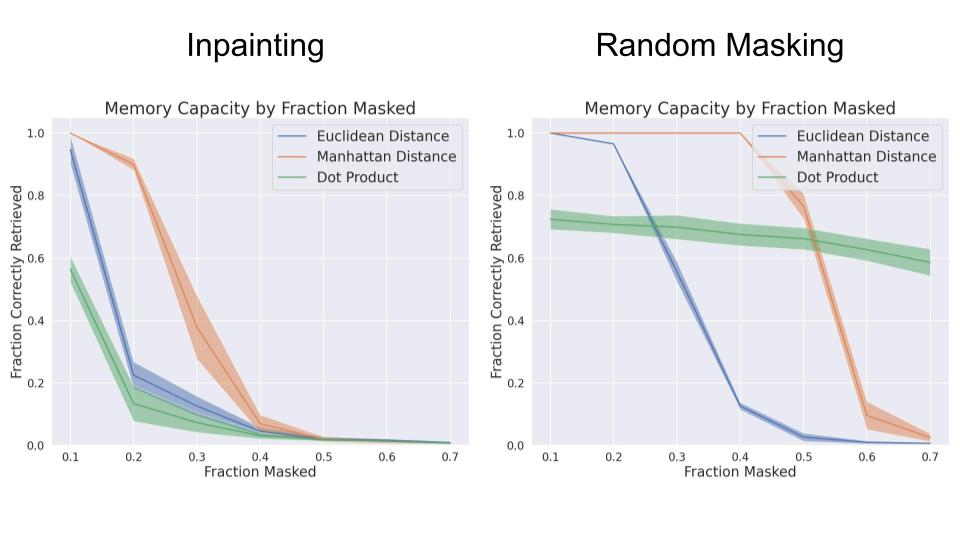}
    \caption{Capacity of networks by similarity function on the two additional perturbations, inpainting and random masking. Error bars are standard deviations across 10 runs.}
    \label{additional_perturbation_figure}
\end{figure}

We see that inpainting followed the standard pattern of superiority of the Manhattan distance metric. However, random masking showed a significantly more robust performance at high capacity for the dot product similarity. This highlights how the optimal similarity function is highly situational depending significantly on the likely set of perturbations that need to be corrected.

\end{document}